%% file: graphformer.tex
\documentclass[10pt,twocolumn,letterpaper]{article}

\usepackage{cvpr}   
\usepackage[accsupp]{axessibility}  

\usepackage{multirow}
\usepackage[table]{xcolor}
\usepackage{amsmath}
\DeclareMathOperator*{\argmax}{arg\,max}
\input{preamble}
\definecolor{cvprblue}{rgb}{0.21,0.49,0.74}
\usepackage[pagebackref,breaklinks,colorlinks,allcolors=cvprblue]{hyperref}

\usepackage[utf8]{inputenc}
\usepackage{booktabs}
\usepackage{siunitx}
\def\paperID{37916} 
\def\confName{CVPR}
\def\confYear{2026}

\title{GraPHFormer: A Multimodal Graph Persistent Homology Transformer for the Analysis of Neuroscience Morphologies}

\author{
Uzair Shah$^{1*}$, Marco Agus$^{1}$\thanks{Equal contribution. Corresponding author: magus@hbku.edu.qa}, Mahmoud Gamal$^{1}$, 
Mahmood Alzubaidi$^{1}$, Corrado Cal\'i$^{2,5,6}$,
\and
Pierre J. Magistretti$^{3}$, 
Abdesselam Bouzerdoum$^{1,4}$, Mowafa Househ$^{1}$\\[2mm]
$^{1}$Hamad Bin Khalifa University, Qatar,  $^{2}$University of Turin, Italy 
$^{3}$BESE, King Abdullah University \\
of Science and Technology, Saudi Arabia,
$^{4}$University of Wollongong, Australia, \\
$^5$Neuroscience Institute  
Cavalieri 
Ottolenghi, Italy, 
$^6$Université Grenoble-Alpes, France
}

\begin{document}
\maketitle

\input{sec/0_abstract}  
\input{sec/intro_up}
\input{sec/2_related}
\input{sec/methods}
\input{sec/results}
\input{sec/5_ablation}

\input{sec/6_conclusion}
\input{sec/acknowledgment}
{
    \small
    \bibliographystyle{ieeenat_fullname}
    \bibliography{graphformer}
}

\input{sec/X_suppl}

\end{document}

%% file: sec/0_abstract.tex
\begin{abstract}
Neuronal morphology encodes critical information about circuit function, development, and disease, yet current methods analyze topology or graph structure in isolation. We introduce GraPHFormer, a multimodal architecture that unifies these complementary views through CLIP-style contrastive learning. Our vision branch processes a novel three-channel persistence image encoding unweighted, persistence-weighted, and radius-weighted topological densities via DINOv2-ViT-S. In parallel, a TreeLSTM encoder captures geometric and radial attributes from skeleton graphs. Both project to a shared embedding space trained with symmetric InfoNCE loss, augmented by persistence-space transformations that preserve topological semantics. Evaluated on six benchmarks (BIL-6, ACT-4, JML-4, N7, M1-Cell, M1-REG) spanning self-supervised and supervised settings, GraPHFormer achieves state-of-the-art performance on five benchmarks, significantly outperforming topology-only, graph-only, and morphometrics baselines. We demonstrate practical utility by discriminating glial morphologies across cortical regions and species, and detecting signatures of developmental and degenerative processes. Code: \url{https://github.com/Uzshah/GraPHFormer}
\end{abstract}

%% file: sec/intro_up.tex
\section{Introduction}
\label{sec:intro}

Neural cell morphology encodes fundamental constraints on information processing, circuit formation, and pathology. Skeletonized reconstructions of neurons and glia enable systematic study of branching patterns, path lengths, tapering, and spatial organization, with direct implications for understanding neurodevelopment, synaptic integration, and neurodegenerative disease. Large public repositories like \textit{NeuroMorpho.Org} provide tens of thousands of digital reconstructions, enabling reproducible benchmarks at scale and driving the need for robust, data-driven representations that capture both topological structure and geometric properties.
\begin{figure}[htb!]
    \centering
    \includegraphics[width=0.24\linewidth]{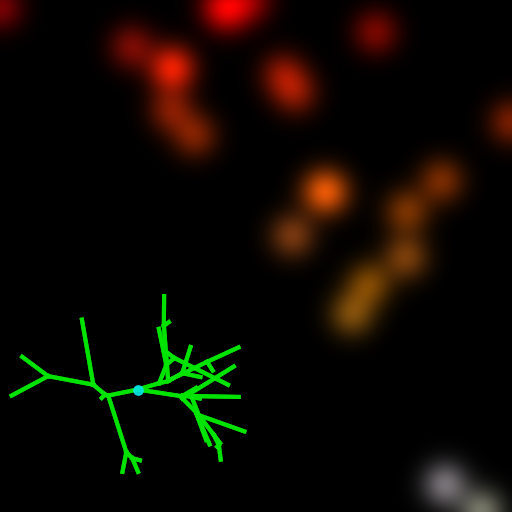}
    \includegraphics[width=0.24\linewidth]{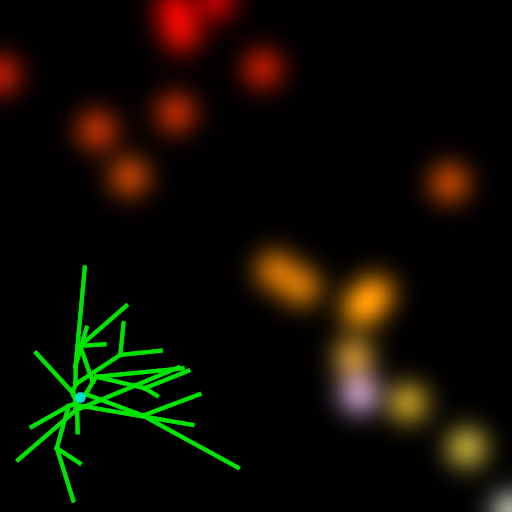}
    \includegraphics[width=0.24\linewidth]{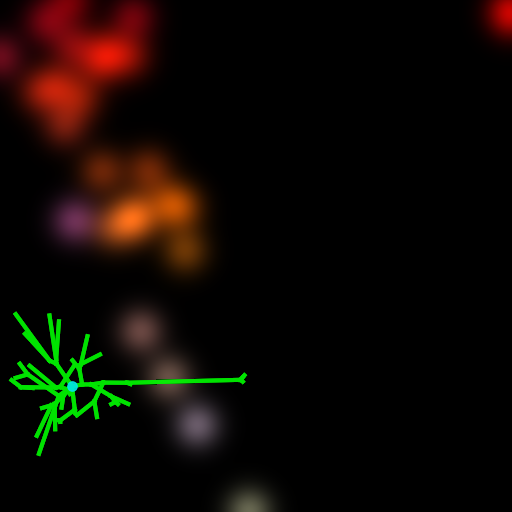}
    \includegraphics[width=0.24\linewidth]{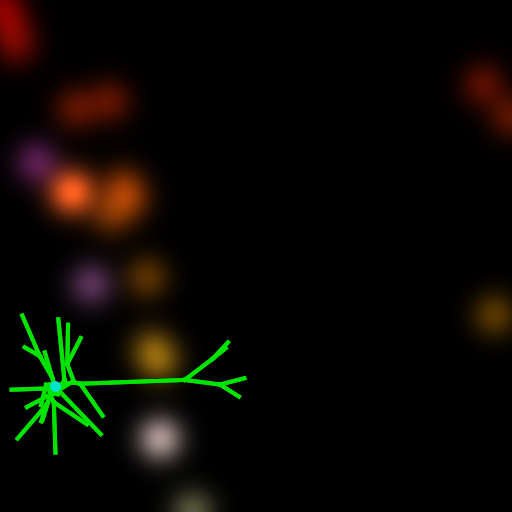}
    \caption{Representative examples showing tree graphs overlaid on their persistence images across different morphological classes.}
    \label{fig:teaser}
\end{figure}
Recent advances in learning on graph-structured morphologies have progressed from graph convolutional networks to graph Transformers with global self-attention and flexible positional encodings~\cite{chen:2022:treemoco,weis:2023:graphdino,sheng:2025:sgtmorph}. In parallel, topological data analysis (TDA) via persistent homology offers provably stable shape summaries. The Topological Morphology Descriptor (TMD) tracks birth and death of branches along filtrations, yielding persistence diagrams vectorizable as images for classification~\cite{kanari2018topological}. However, existing methods treat these views in isolation: graph-based approaches encode branching patterns but miss global geometric invariants, while topology-based methods capture spatial arrangements but lose fine-grained structural information. This fundamental limitation motivates a multimodal approach that exploits complementary strengths.
%

We propose \textbf{GraPHFormer}, a multimodal architecture that unifies (i) a \emph{vision} stream operating on multi-channel persistence images derived from the skeleton and (ii) a \emph{graph} stream processing the original morphological tree with geometric and radial attributes. We adapt the CLIP framework~\cite{clipradford21a} to neuronal morphology, training dual encoders—a vision Transformer (DinoV2) for persistence images and a graph encoder (TreeLSTM) for tree structures—to project into a shared embedding space using symmetric InfoNCE loss. This joint learning leverages topology for global branching motifs and graphs for local spatial detail.

In summary, we provide the following contributions:
\begin{itemize}
  \item \textbf{Multi-channel persistence images.} We introduce a three-channel RGB persistence image encoding complementary morphological aspects: (R) unweighted TMD-style density preserving spatial structure, (G) persistence-weighted density emphasizing topologically salient features, and (B) radius-weighted density capturing radial characteristics ( Sec.~\ref{sec:persistence_generation}). 
  
  \item \textbf{Persistence space augmentation.} We propose augmentation strategies applied directly in topological feature space before image generation—including birth-death jittering, persistence scaling and radius perturbation ( Sec.~\ref{sec:per_space_aug}). 
  
  \item \textbf{Multimodal contrastive learning.} We adapt CLIP to neuron representation learning, to our knowledge the first such application, training dual encoders to align tree graphs and persistence images in a shared embedding space. This enables joint learning of complementary structural and topological features (Sec.~\ref{sec:method}).
  
\end{itemize}
%

GraPHFormer consistently and significantly outperforms state-of-the-art topology-only, graph-only, and morphometrics baselines. In self-supervised learning, we achieve substantial improvements: 86.2\% on BIL-6 (+4.9 over SGTMorph), 72.7\% on JML-4 (+6.1), and 83.8\% on N7 (+4.0), with remarkably low variance (±0.6-4\%). Under supervised learning, we establish new state-of-the-art results on five of six benchmarks, reaching 93.51\% on BIL-6 and 92.3\% on N7. Beyond accuracy metrics, our framework demonstrates practical relevance for discriminating neuronal and glial morphologies across cortical areas, species, developmental trajectories, and degenerative conditions, providing neuroscientists with robust tools for quantitative morphology analysis.

%% file: sec/2_related.tex
\section{Related work}
\label{sec:related}

Our work deals with deep learning methods for graph-based neuroscience morphologies, 
topology-based descriptors for branched morphologies, and multi-modal graph transformers. We do not aim to provide here an extensive overview of the literature corpus in the field: we refer interested readers to 
the recent surveys related to graph learning~\cite{khoshraftar:2024:survey_grl,yang:2024:survey_gl} and to the usage of topology data analysis in learning~\cite{zia:2024:topological} and graph neural networks~\cite{pham:2025:tda_gnn_survey}. In the following, we discuss the methods most closely related to our work.

\paragraph{Deep learning for graph-based neuroscience morphology.}
Learning on irregular neuronal and glial skeletons has progressed from early image– and sequence–based encodings to point– and graph–native architectures. Image projections of 3D trees with stacked convolutional autoencoders provided the first learning–based neural morphometrics, trading geometric fidelity for CNN compatibility~\cite{zhang:2021:neuron}. Sequence-aware models such as TRNN integrated ordered node information with image features to better capture spatiotemporal dependencies along arbors~\cite{fan:2024:morphrep}. Point-cloud approaches (e.g., MorphoGNN) leveraged dynamic graph construction to learn low-dimensional but discriminative latent spaces while preserving local geometry~\cite{zhu:2023:morphognn,troidl2025global}. Hybrid designs fused PointNet-style morphological descriptors with circuit connectivity in GNNs to jointly represent shape and network context~\cite{zhang:2021:neuron}. Because gold-standard labels are scarce and heterogeneous across labs and species, self-/contrastive learning became central: MorphVAE learned branch-level latent codes via variational sequence models but only weakly captured inter-branch structure~\cite{laturnus2021morphvae}; MACGNN applied graph contrastive objectives to neuronal trees~\cite{zhao:2024:MACGNN}; and graph-level self-distillation and momentum-contrast variants further improved label efficiency on morphology classification~\cite{kreuzer:2021:rethinking,rong:2020:self}. Our method applies contrastive learning on a multimodal graph transformer exploiting  state of the art GNNs~\cite{chen:2022:treemoco,weis:2023:graphdino}, enriched with persistent homology image representations.

\paragraph{Graph Transformers and multimodal graph learning.}
Transformer models for graphs combine global self-attention with structural priors to overcome the locality limits of message passing. Spectral positional encodings based on Laplacian eigenvectors and full-spectrum variants inject global structure into token embeddings~\cite{wu:2021:representing,kreuzer:2021:rethinking}; attention-bias formulations incorporate distances, higher-order walk information, and edge attributes to inform pairwise attention~\cite{wu:2022:nodeformer,rong:2020:self}. These advances enable long-range dependency modeling across complex arbors and are increasingly adopted in neuroscience morphology pipelines~\cite{fan:2024:morphrep,zhang:2021:neuron}.
Beyond single-modality graphs, \emph{multimodal} graph Transformers align graph tokens with auxiliary modalities (e.g., images, sequences, curated descriptors) via cross-attention or token grafting, improving robustness when labels are scarce or distributions shift~\cite{tang:2023:scmoformer, wu:2022:nodeformer}. 
SGTMorph~\cite{sheng:2025:sgtmorph} exemplifies this hybrid direction for neuronal trees: it couples the \emph{local} topological modeling strengths of GNNs with the \emph{global} relational reasoning capacity of Transformers to explicitly encode neuronal structure; uses a \emph{random-walk positional encoding} to facilitate information propagation over complex arbors; and introduces a \emph{spatially invariant encoding} to improve adaptability across diverse morphologies. Our approach follows this line by pairing a graph Transformer on the original skeleton with a vision Transformer operating on topology-derived images, enabling complementary inductive biases—global branching motifs in an image canvas and fine-grained connectivity/radii on the graph—to be learned jointly. We adapt CLIP \cite{clipradford21a} for this and to our knowledge, this is the first application of CLIP-style contrastive learning to neuron representation learning.

\paragraph{Topological data analysis (TDA) for morphology.}
TDA provides stable, scale-aware summaries of shape and connectivity via persistent homology, which tracks the birth and death of features across filtrations and yields barcodes/diagrams vectorizable as persistence images or learnable embeddings~\cite{edelsbrunner2010computational,adams2017persistence,carriere2020multiparameter}. For neuronal trees, the Topological Morphology Descriptor (TMD) encodes branching organization using soma-centric filtrations (radial or geodesic), producing characteristic distributions linked to development, region, and species~\cite{kanari2018topological,kanari2020objective,kanari2019objective,kanari2025mice}. Topological summaries have proven competitive for cell-type classification and population comparisons when paired with classical ML or shallow CNNs over persistence images~\cite{zia:2024:topological,curto2025topological,aktas2019persistence}. Our work builds on these insights in two ways: (i) we propose a multi-channel persistence image for morphological trees that augments an \emph{unweighted} TMD-style density with branch-length (persistence) and mean-radius channels, and (ii) we fuse this topological view with a graph Transformer over the original skeleton within a multimodal architecture. This design exploits complementary strengths—translation/rotation-agnostic global branching patterns from TDA and locality–aware geometric reasoning from graphs—to robustly advance morphology analysis across datasets and supervision regimes.

%% file: sec/methods.tex
\section{Methodology}
\label{sec:method}
\subsection{Overview}
We propose a multimodal contrastive learning framework that addresses a key limitation in neuron morphology analysis: existing methods rely on single modalities capturing incomplete structural aspects. Graph-based approaches like TreeMoCo \cite{chen:2022:treemoco} and GraphDINO \cite{weis:2023:graphdino} encode branching patterns but miss geometric variations, while image-based methods capture spatial arrangements but lose topological information.
Our key insight is leveraging complementary strengths of both representations through multimodal alignment. We adapt CLIP \cite{clipradford21a} to neuron morphology using: (1) tree graphs encoding hierarchical branching structure, and (2) persistence images capturing topological invariants via TDA. 
Our dual-encoder architecture uses a tree encoder (TreeLSTM or GraphDINO) for directed graphs and an image encoder (DINOv2 \cite{oquab2023dinov2} or ResNet-18 \cite{he2016deep}) for multi-channel persistence images. Both project to a shared n-dimensional space via MLP heads. We maximize agreement between corresponding tree-image pairs while minimizing non-corresponding pairs using symmetric InfoNCE loss, creating a unified embedding space where morphologically similar neurons cluster together.
For evaluation, following TreeMoCo's protocol, we combine learned embeddings through concatenation or addition, then apply a frozen 
k-NN classifier. 

\subsection{Persistence Image Generation}
\label{sec:persistence_generation}
To capture the topological signatures of neuronal structures, we convert each neuron graph into a multi-channel persistence image representation using Topological Morphology Descriptor (TMD) analysis~\cite{kanari2018topological}. This process transforms the discrete tree structure into a continuous topological representation that encodes branching patterns and morphological complexity. Intuitively, TMD sweeps a distance threshold outward from the soma and records where branches \textit{begin} and \textit{end}, capturing 
the tree's shape in a compact descriptor.

\noindent \textbf{Neuron Tree Preprocessing:}
Neuron reconstructions stored in SWC format are parsed to extract seven key attributes for each node: node ID, node type ($t$), 3D coordinates $(x, y, z)$, radius ($r$), and parent ID. Let $\mathcal{V} = \{v_1, v_2, \ldots, v_n\}$ denote the set of nodes and $\mathcal{E}$ denote the parent-child edges. The tree structure $\mathcal{T} = (\mathcal{V}, \mathcal{E})$ is constructed by building a parent-child relationship map: $\mathcal{V} \rightarrow 2^{|\mathcal{V}|}$, where the root node $v_{\text{s}}$ (soma) is identified by parent ID = $-1$.

\noindent \textbf{Radial Distance Filtration:}
We employ radial distance from the soma as the \textit{filtration function}, a scalar value assigned to each node that orders the tree structure for topological analysis, for persistent homology computation.. For each node $v_i \in \mathcal{V}$ with coordinates $(x_i, y_i, z_i)$, we compute the Euclidean distance to the soma at $(x_s, y_s, z_s)$:

\begin{equation}
d_{\text{raw}}(v_i) = \sqrt{(x_i - x_s)^2 + (y_i - y_s)^2 + (z_i - z_s)^2}
\end{equation}

To ensure monotonicity along root-to-leaf paths—a requirement for meaningful persistence computation—we enforce a cumulative maximum constraint via breadth-first traversal from the root:

\begin{equation}
f(v_i) = \begin{cases}
0 & \text{if } v_i = v_s \\
\max\{f(\text{parent}(v_i)), d_{\text{raw}}(v_i)\} & \text{otherwise}
\end{cases}
\end{equation}

This ensures that the filtration function $f: \mathcal{V} \rightarrow \mathbb{R}^+$ is non-decreasing along any path from the soma to dendritic terminals, satisfying the requirement that $f(v_i) \leq f(v_j)$ whenever $v_i$ is an ancestor of $v_j$.

\noindent \textbf{Persistence Pair Computation:}
Persistence pairs are computed using the TMD elder-rule algorithm~\cite{kanari2018topological}, which identifies topological features (branches) as $(b_i, d_i)$ pairs, where $b_i$ (birth) is the distance at which a branch tip appears and $d_i$ (death) is the distance at which it merges into a larger branch.. The algorithm processes nodes in post-order traversal and maintains a champion for each subtree—defined as the leaf node with maximum filtration value. 

For each internal node $v$ with children $\{c_1, c_2, \ldots, c_k\}$, let $(\ell_{j^*}, f(\ell_{j^*}))$ denote the champion (leaf node, filtration value) of the subtree rooted at $c_j$. The champion of $v$ is defined as:

\begin{equation}
\chi(v) = (\ell_{j^*}, f(\ell_{j^*})), \quad \text{where } j^* = \argmax_{j \in \{1,\ldots,k\}} f(\ell_j)
\end{equation}

For all non-champion children $c_j$ where $j \neq j^*$, a persistence pair is created:

\begin{equation}
\tau_j = (b_j, d_j, \ell_j,v), \quad b_j = f(\ell_j), \quad d_j = f(v)
\end{equation}

where $b_j$ is the \textit{birth value} (filtration value at the leaf), $d_j$ is the \textit{death value} (filtration value at the bifurcation), $\ell_j$ is the leaf node identifier, and $v$ is the death node identifier. The set of all persistence pairs is denoted as $\mathcal{P} = \{(b_i, d_i, \ell_i, v_i)\}_{i=1}^m$.

\noindent \textbf{Feature Enrichment:}
For each persistence pair $(b_i, d_i, \ell_i, v_i) \in \mathcal{P}$, we compute additional geometric features by tracing the path $\pi_i = \text{path}(\ell_i, v_i)$ from the leaf node to the bifurcation node:

\begin{equation}
\delta_i = b_i - d_i, \bar{r}_i = \frac{1}{|\pi_i|} \sum_{v \in \pi_i} r(v)
\end{equation}


where $r(v)$ denotes the radius attribute of node $v$. These features provide complementary information about branch geometry beyond pure topological persistence. Each pair is thus represented as a feature vector:

\begin{equation}
  \rho_i = (b_i,\, d_i,\, \delta_i,\, \bar{r}_i)
\end{equation}

\noindent \textbf{Persistence Image Construction:}
We construct a three-channel RGB persistence image $\mathbf{I} \in \mathbb{R}^{H \times W \times 3}$ (wherever not indicated we consider $H = W = 112$), each channel encodes different aspects of neuronal morphology through weighted density estimation in the 2D $(b, p)$ space, analogous to a scatter plot of branch features, where $b$ is the birth distance and $p = b - d$ is the branch lifetime (persistence).

\begin{figure}
    \centering
    \includegraphics[width=0.95\linewidth]{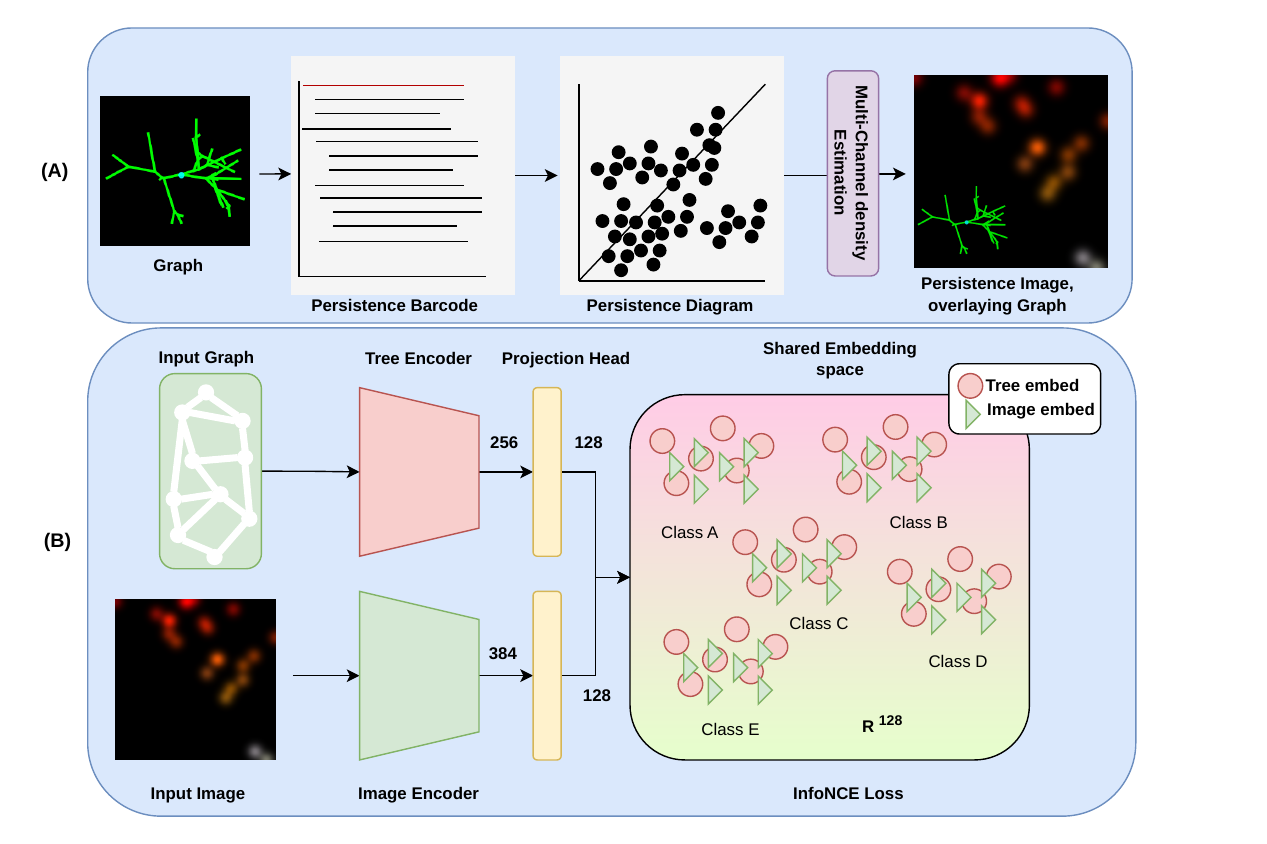}
    \caption{Overview of the proposed dual-modality framework.
\textbf{(A)} Neuronal morphologies are converted into multi-channel RGB persistence images via radial distance filtration, TMD-based persistence pair extraction, feature enrichment, and Gaussian kernel density estimation.
\textbf{(B)} A self-supervised architecture jointly trains a tree encoder and an image encoder, projecting both modalities into a shared embedding space. Alignment between graph and image features is optimized via contrastive (InfoNCE) loss following the CLIP paradigm.}
    \label{fig:pipeline}
\end{figure}

\noindent \textbf{Multi-Channel Density Estimation:}
For each channel $c \in \{R, G, B\}$, we compute a weighted density map 
using Gaussian kernels. Let $w_i^{(c)}$ denote the channel-specific weight 
for pair $i$, and let $(x, y)$ represent the 2D coordinate grid of the 
persistence diagram, where the axes correspond to birth time and persistence 
values, respectively. The channel $c$ is computed as:

\begin{equation}
\small
\begin{aligned}
w_i^{(R)} &= 1, \quad w_i^{(G)} = \delta_i, \quad w_i^{(B)} = \bar{r}_i \\
\mathbf{I}^{(c)}(x, y) &= \sum_{i=1}^{m} w_i^{(c)} \cdot \frac{1}{2\pi\sigma^2} \exp\!\left(-\frac{(x-b_i)^2+(y-p_i)^2}{2\sigma^2}\right)
\end{aligned}
\end{equation}

where $\sigma$ pixels controls the kernel bandwidth (wherever not indicated 
we use $\sigma=16$). $b_i$ and $p_i$ denote the birth time and persistence 
value of the $i$-th topological feature (persistence pair), respectively, 
which serve as the 2D coordinates in the persistence diagram space.







\noindent \textbf{Persistence Space Augmentation:}
\label{sec:per_space_aug}
Standard image-space augmentations (e.g., rotation, scaling, color jittering) are ineffective for persistence images because they violate the semantic correspondence between pixel positions and topological features. For instance, rotating a persistence image arbitrarily reassigns the meaning of the birth-persistence axes, corrupting the topological information. Similarly, photometric augmentations disrupt the carefully calibrated density weights that encode geometric properties.

To address this limitation, we propose \textit{persistence space augmentation}, which applies transformations directly to the feature vectors $\{\mathbf{p}_i\}$ before image generation, preserving the semantic structure of the topological representation.

\noindent \textbf{Augmentation Operations:}
During training, we stochastically apply the following transformations to each feature vector $\rho_i = (b_i, d_i, \delta_i, \bar{r}_i)$:

\begin{enumerate}
\item \textbf{Birth-Death Jittering:} Add Gaussian noise to birth and death values:
\begin{equation}
b_i' = b_i + \mathcal{N}(0, \sigma_b^2), \quad d_i' = d_i + \mathcal{N}(0, \sigma_d^2)
\end{equation}
where $\sigma_b, \sigma_d \in [0.01, 0.05] \times (b_{\max} - b_{\min})$ are sampled uniformly.

\item \textbf{Persistence Scaling:} Scale persistence values to simulate variations in branch extent:
\begin{equation}
\delta_i' = \delta_i \cdot \alpha, \quad \alpha \sim \mathcal{U}(0.9, 1.1)
\end{equation}

\item \textbf{Radius Perturbation:} Perturb mean radius to model reconstruction uncertainty:
\begin{equation}
\bar{r}_i' = \bar{r}_i \cdot \beta, \quad \beta \sim \mathcal{U}(0.85, 1.15)
\end{equation}

\end{enumerate}

After augmentation, the modified feature vectors $\{\mathbf{p}_i'\}$ are used to generate the persistence image, ensuring that augmentations respect the topological semantics.

\subsection{Encoder Architecture}
\paragraph{TreeEncoder} 
Neuron morphology in SWC format defines each node by $(id, type, x, y, z, r, parent)$. We construct directed tree $\mathcal{T} = (\mathcal{V}, \mathcal{E})$ where vertices represent nodes and edges encode parent-child relationships. Each node $v$ has feature vector $\mathbf{x}_v \in \mathbb{R}^5$ containing spatial coordinates $(x, y, z)$, radius $r$, and path length from soma. Features are z-score normalized. We employ TreeLSTM \cite{chen:2022:treemoco} for hierarchical aggregation. Given node $v$ with children $C(v)$:
$$\mathbf{h}_v = \text{TreeLSTM}(\mathbf{x}_v, \{\mathbf{h}_c\}_{c \in C(v)})$$
where $\mathbf{h}_v \in \mathbb{R}^{D}$ is the hidden state. Child states are aggregated via sum pooling. Bottom-up traversal from leaves to root yields root embedding $\mathbf{h}_\text{root}$ as the tree representation.

\paragraph{Image Encoder} 
Each neuron converts to 2D persistence image $\mathcal{I} \in \mathbb{R}^{H \times W \times 3}$ through TDA (detailed in Section \ref{sec:persistence_generation}). Three RGB channels encode: (R) unweighted feature density, (G) persistence-weighted density, (B) mean-radius-weighted density. This multi-scale encoding captures both local branch complexity and global patterns. We employ DINOv2-ViT-S/14 \cite{oquab2023dinov2} as image encoder. The architecture has patch embedding (14×14 patches), 12 transformer blocks with 6 attention heads, and hidden dimension 384. We fine-tune the entire backbone to adapt pre-trained natural image representations to the topological domain.

\paragraph{Projection Heads}
Following SimCLR \cite{chen2020simple}, we employ 2-layer MLP projection heads for both encoders to map their outputs to a shared D-dimensional embedding space where contrastive learning occurs. The projection heads are defined as:
\begin{align*}
\text{$g_\text{tree}$:} \quad &\mathbb{R}^{256} \xrightarrow{\text{Linear}} \mathbb{R}^{256} \xrightarrow{\text{Linear}} \mathbb{R}^{128} \\
\text{$g_\text{image}$:} \quad &\mathbb{R}^{384} \xrightarrow{\text{Linear}} \mathbb{R}^{256} \xrightarrow{\text{Linear}} \mathbb{R}^{128}
\end{align*}
During training, embeddings are $\ell_2$-normalized before computing similarities. 

\subsection{Self-Supervised Learning Strategy}

The vast majority of morphological datasets lack annotations, as manual 
labeling demands significant expertise, making it economically unfeasible 
at scale. Self-supervised learning addresses this by extracting meaningful 
representations from unlabeled data without expensive annotation.

Consider a collection of $N$ neurons, where each neuron $n_i$ is characterized through two distinct but complementary representations: a tree graph $\mathcal{T}_i$ that captures the hierarchical organization of neuronal branches, and a persistence image $\mathcal{I}_i$ that encodes topological properties. We aim to train encoders $f_\text{tree}$ and $f_\text{image}$ that project these disparate modalities into a unified embedding space, where representations of the same neuron align across modalities, facilitating cross-modal matching and enabling various downstream applications without requiring labels.

We adopt the symmetric InfoNCE formulation from CLIP \cite{clipradford21a}, which enables mutual supervision between modalities. Within each training batch of $N$ neurons, we compute $\ell_2$-normalized embeddings $\mathbf{z}_i^t = g_\text{tree}(f_\text{tree}(\mathcal{T}_i))$ and $\mathbf{z}_i^v = g_\text{image}(f_\text{image}(\mathcal{I}_i))$ where 
$g_\text{tree}$ and $g_\text{image}$ are modality-specific projection heads 
that map encoder outputs into the shared embedding space.. Pairwise similarity is measured as $s_{ij} = \langle \mathbf{z}_i^t, \mathbf{z}_j^v \rangle / \tau$, where the learnable temperature $\tau$ modulates the sharpness of the similarity distribution. The training objective combines two symmetric cross-entropy terms:
\begin{equation}
\mathcal{L} = -\frac{1}{2N} \sum_{i=1}^N \left[ \log \frac{\exp(s_{ii})}{\sum_{j=1}^N \exp(s_{ij})} + \log \frac{\exp(s_{ii})}{\sum_{j=1}^N \exp(s_{ji})} \right]
\end{equation}
This formulation enforces bidirectional consistency by encouraging high 
similarity between matched tree-image pairs (positives) while suppressing 
similarity with unmatched pairs (negatives). Compared to approaches 
requiring explicit negative mining~\cite{hao2024towards} or momentum-based 
memory queues~\cite{sheng:2025:moq}, our method requires no additional memory structures or sampling heuristics,
as within-batch negatives scale as $\mathcal{O}(N)$ with batch size, 
compared to $\mathcal{O}(K)$ memory overhead for queue-based 
approaches~\cite{chen2020simple} and $\mathcal{O}(N^2)$ pairwise 
computations for explicit mining.

\subsection{Supervised Learning Strategy}
For downstream tasks, we remove the projection heads from both pretrained encoders, fuse their outputs via concatenation, and attach a classification head. Fine-tuning follows a two-stage approach: linear probing of the classification head alone, followed by end-to-end fine-tuning of the complete model. 

\begin{table*}[htb]
\centering
\caption{Performance comparison of GraPHFormer against state-of-the-art methods across six neuronal morphology datasets (classification accuracy in \%). \textbf{SL}: Supervised Learning with full labels. \textbf{SS}: Self-Supervised Learning with frozen kNN evaluation following TreeMoCo~\cite{chen:2022:treemoco}. Bold: best performance; underline: second-best. SS methods show mean ± std over five seeds. Missing entries (-) indicate unreported results. Datasets: BIL-6, ACT-4, JML-4 (4 cell types), Neuron7 (N7), M1-Cell (cell types), M1-REG (regions). Baselines from MorphRep~\cite{fan:2024:morphrep} (M1-*) and SGTMorph~\cite{sheng:2025:sgtmorph} (others). TreeMoCo* denotes our fine-tuned tree encoder.}
\begin{tabular}{llcccccc}

\hline
& Method & BIL-6 & ACT-4 & JML-4 & N7 & M1-Cell & M1-REG \\
\hline
\multirow{7}{*}{SL} 
& MorphVAE & 66.80 & 41.05 & 40.00 & 73.73 & 60 & 64 \\
& TRNN & 76.56 & 46.32 & 51.43 & - & -&- \\
& Morphometrics & 81.27 & 52.08 & 52.58 & 82.45 & -&- \\
& MorphoGNN & 76.57 & 54.16 & 65.83 & 85.58 &- &- \\
& TreeMoCo & 85.55 & 62.11 & 64.29 & - &- &- \\
& MorphRep & 88.0 & \underline{77.0} & - & - & 72.0 & \underline{76.0} \\
& SGTMorph & \underline{88.9} & \textbf{79.3} & \underline{72.4} & \underline{89.0} & -& -\\
\cline{2-8}
\rowcolor{green!15} & \textbf{TreeMoco*} & 88.3 & 53.5 & 63.7 & 82.5 & \underline{74.7} & 71.6\\
\rowcolor{green!15} & \textbf{PI} & 80.5 & 54.1 & 69.1 & 84.7 & 73.5 & 74.1\\
\rowcolor{green!15} & \textbf{GraPHFormer} & \textbf{93.51} & 65.5 & \textbf{76.5} & \textbf{92.3} & \textbf{76.5} & \textbf{79.1}\\
\hline
\multirow{6}{*}{SS}
& GraphCL & 66.3 & 53.9 & 59.7 & - & - & -\\
& TreeMoCo & 76.5 & 56.8 & 62.7 & 71.8& -& -\\
& MACGNN & 69.5 & 45.9 & 61.4 & 71.1 & -& -\\
& MorphRep & 81.0 & $\mathbf{66.0}$ & - & - & \underline{73.0} & \underline{75.0} \\
& GraphDINO & $79 \pm 1$ & $54 \pm 5$ & $63 \pm 6$  & 68.3 & 68 & 71\\
& SGTMorph & \underline{$81.3 \pm 1$} & \underline{63.2$\pm 3$} & \underline{$66.6 \pm 5$}  & \underline{79.8 $\pm 0.8$} & - & -\\
\cline{2-8}
\rowcolor{green!15} & \textbf{GraPHFormer} & $\mathbf{86.2 \pm 1}$ & 59.1 $\pm 4$ & $\mathbf{72.7}\pm 3$ & $\mathbf{83.8}\pm 0.6$ & $\mathbf{75.5} \pm 1.7$& $\mathbf{76.5} \pm 1$\\
\hline

\end{tabular}
\label{tab:results}
\end{table*}

%% file: sec/results.tex
\section{Experiments}
\subsection{Dataset and Experimental Settings.}
We follow the data-splitting strategy of TreeMoCo~\cite{chen:2022:treemoco} with a 70\%–30\% train–test split, consistent with GraphDINO~\cite{weis:2023:graphdino} and SGTMorph~\cite{sheng:2025:sgtmorph}, for fair comparison. 

\noindent \textbf{Datasets.} 
We evaluate our framework on five neuronal morphology datasets:

\begin{itemize}[leftmargin=1.5em]
    \item \textbf{N7 (Neuro7)}~\cite{zhu2023data, ascoli2007neuromorpho}: Seven representative mouse neuron types from NeuroMorpho.Org.
    \item \textbf{ACT (Allen Cell Types)}~\cite{gouwens2019classification}: Fully reconstructed neurons from the mouse visual cortex.
    \item \textbf{BIL (BICCN fMOST)}~\cite{peng2021morphological}: Rat neurons from cortex, claustrum, striatum, and thalamus.
    \item \textbf{JML (Janelia MouseLight)}~\cite{gao2023single}: Long-range projection neurons across thalamus, hippocampus, cortex, and hypothalamus.
    \item \textbf{M1-EXC}~\cite{fan:2024:morphrep, scala2021phenotypic}: Neurons with dual annotations for cell type and projection pattern, enabling multi-task analysis.
\end{itemize}
In the supplementary material we showcase some examples of potential applications for glia discriminations on morphologies extracted from NeuroMorph.org.

\noindent \textbf{Experimental Settings.} 
GraPHFormer was implemented in PyTorch and trained on an NVIDIA RTX 4090. 
It features a dual-encoder design—TreeLSTM for tree structures and DinoV2-small for persistence images—with dropout ($p=0.1$) applied after each transformer layer. 
Optimization used AdamW ($\beta_1{=}0.9$, $\beta_2{=}0.999$, $\lambda{=}0.05$), a learning rate of $5\times10^{-4}$, 20-epoch warmup, and cosine annealing over 300 epochs (batch size 128). 
For downstream evaluation, we adopted the TreeMoCo protocol using a frozen $k$-NN classifier ($k{=}20$, except $k{=}5$ for JML), reporting the best results averaged over five random seeds.

\subsection{Results}
\label{sec:results}
Table~\ref{tab:results} presents performance comparisons across six neuronal morphology benchmarks under supervised (SL) and self-supervised (SS) learning. For SL, we finetune from self-supervised pretrained weights. For SS, we use frozen kNN evaluation ($k=20$, except $k=5$ for JML-4 ) following TreeMoCo~\cite{chen:2022:treemoco}. All SS results show mean ± std over five seeds.

\paragraph{Self-Supervised Learning}

GraPHFormer achieves state-of-the-art on five of six benchmarks. On \textbf{BIL-6}, we reach \textbf{86.2 ± 1\%}, surpassing SGTMorph (81.3\%) and MorphRep (81.0\%) by ~5 points despite using only benchmark data versus MorphRep's large-scale pretraining. On \textbf{N7}, we achieve \textbf{83.8 ± 0.6\%} with isolated training, outperforming SGTMorph (79.8\%) by 4.0 points with exceptionally low variance.

Suprisingly, \textbf{ACT-4} proves challenging at 59.1 ± 4\%, trailing MorphRep (66.0\%) and SGTMorph (63.2\%). We hypothesize cortical layer classification requires contextual features beyond single-cell morphology, and that competing baselines used additional training data covering the same region.

\paragraph{Supervised Learning}

To validate our multimodal fusion, we include \textbf{PI} (DinoV2s)—an image-only baseline using persistence images alone.  PIachieves moderate performance (80.5\% on BIL-6, 69.1\% on JML-4 ), consistently underperforming GraPHFormer by 13.0 and 7.4 points respectively. This demonstrates that persistence images alone are insufficient and that graph-based structural features are essential for discriminative morphology representations.

Our full \textbf{GraPHFormer} with multimodal fusion reaches \textbf{93.51\%} on BIL-6, surpassing SGTMorph (88.9\%) by 4.6 points, and \textbf{76.5\%} on JML-4 , exceeding SGTMorph (72.4\%) by 4.1 points. The substantial gap between DinoV2s and GraPHFormer validates the complementarity of topological and graph representations.

\textbf{ACT-4} remains problematic at 65.5\%, trailing SGTMorph (79.3\%) and MorphRep (77.0\%) by 13.8 and 11.5 points. ACT-4 neuron classes depend critically on absolute cortical depth and laminar position---discriminative cues our translation-invariant TMD and graph encoders do not capture. In contrast, SGTMorph retains geometric and coordinate-dependent features, while MorphRep's large-scale pretraining implicitly encodes regional anatomical patterns. Notably,  PIachieves only 54.1\%, indicating persistence images particularly struggle with layer classification. The persistent gap even under full supervision confirms that layer classification requires features beyond our topology-structure fusion---specifically spatial coordinates, regional context, or circuit-level connectivity that our architecture deliberately abstracts away. Overall, GraPHFormer achieves state-of-the-art or competitive performance on five of six benchmarks, with the image-only baseline validating the necessity of multimodal fusion.

\noindent \textbf{Cross-dataset evaluation} To assess the generalizability of our method, we trained GraPHFormer on one dataset using self-supervised learning for 100 epochs and evaluated its performance on other datasets. Results are presented in Table~\ref{tab:generalizability}. We examined four scenarios: (1) training on Joint (ACT, BIL, and JML) and evaluating on Neuron7, M1-EXC-Cell, and M1-EXC-Region; (2) training on Neuron7 and evaluating on other datasets; (3) training on M1-EXC and evaluating on other datasets; and (4) training on all five datasets combined.
Models trained on the All-dataset configuration achieved the highest performance across most evaluation datasets, reaching 85.71\% on BIL-6 and 84.45\% on N7. The Neuron7-trained model demonstrated notable cross-dataset transferability, achieving 84.42\% on BIL-6 and 74.07\% on M1-Reg. However, transfer to ACT-4 remained challenging, with accuracies around 50-55\% across all training configurations.

\begin{table}[h]
\centering
\small
\caption{Cross-dataset generalizability evaluation. Models trained on datasets shown in columns (top row) are evaluated on datasets shown in rows (first column). Joint combines ACT, BIL, and JML; All-dataset includes all five datasets. All models trained for 100 epochs.}
\label{tab:generalizability}
\begin{tabular}{lcccc}
\toprule
\textbf{Dataset} & \textbf{Joint} & \textbf{Neuron7} & \textbf{M1-EXC} & \textbf{All-dataset} \\
\midrule
BIL-6 & - & 84.42 & 76.62 & 85.71 \\
JML-4  & - & 72.57 & 56.64 & 76.99 \\
ACT-4 & - & 52.08 & 50.00 & 54.86 \\
N7 & 80.86 & - & 72.01 & 84.45 \\
M1-Cell & 71.08 & 69.88 & - & 73.49 \\
M1-Reg & 62.96 & 74.07 & - & 75.31 \\
\bottomrule
\end{tabular}
\end{table}

\noindent\textbf{Cross-Domain Generalizability.} To assess whether GraPHFormer generalizes beyond neurons, we conduct cross-domain transfer experiments with glial morphologies from NeuroMorpho.Org~\cite{ascoli2007neuromorpho} (11,925 samples across four species). Training on neurons and testing on glia yields 78.87\% species classification accuracy, while the reverse transfer, glia-trained model evaluated on neuronal benchmarks, achieves competitive performance on N7 (82.78\%) and BIL-4 (81.82\%), approaching within-domain results. This suggests that GraPHFormer captures morphological signatures shared across cell types, with full details in supplementary material.

\subsection{Complementarity Analysis}

We extracted learned embeddings from the tree encoder (TreeLSTM-Double, 256-dim)
and image encoder (DINOv2-ViT-S/14, 384-dim) of trained supervised models and
evaluated complementarity using k-nearest neighbors (k=20) classification.

We quantified complementarity through multiple metrics: (1) \textbf{Pearson correlation} measures feature-wise linear correlation after dimension alignment via PCA,
with values near zero indicating different encoded information; (2) \textbf{Distance
correlation (RSA)} uses Spearman's on pairwise distance matrices to assess
geometric structure similarity; 
(3) \textbf{Complementarity score} is the percentage of samples where exactly one
modality is correct while the other fails; (4) \textbf{Fusion rescue rate} measures
the percentage of hard cases (both modalities wrong) where fusion succeeds, indicating
synergistic benefits.

Analysis on Neuron7 (418 test samples, 7 classes) and JML-4 (113 test samples, 4 classes)
revealed low cross-modal correlation (Pearson $ = 0.040$ and $0.083$, RSA
$ = 0.080$ and $0.060$, both $p < 0.001$), indicating tree and image modalities
capture different aspects of neuronal morphology (Table~\ref{tab:complementarity_comparison}).
Both datasets benefited from fusion: Neuron7 improved from 81.8\% to 83.5\% (+1.7\%),
while JML-4 improved from 64.6\% to 66.4\% (+1.8\% with weighted fusion). Despite similar
fusion gains, JML-4 showed higher complementarity score (52.2\% vs 38.0\%) and better
rescue rate (21.1\% vs 17.3\%), suggesting the modalities capture more distinct
discriminative features in the JML-4 dataset.

\begin{table}[htbp]
\centering
\caption{Complementarity analysis comparing Neuron7 and JML-4  datasets on test sets.}
\label{tab:complementarity_comparison}
\small
\begin{tabular}{llcc}
\toprule
\textbf{Category} & \textbf{Metric} & \textbf{Neuron7} & \textbf{JML-4 } \\
\midrule
Dataset Info
  & Samples & 418 & 113 \\
  & Classes & 7 & 4 \\
\midrule
\multirow{2}{*}{Correlation}
  & Pearson  & 0.040 & 0.083 \\
  & RSA  & 0.080\textsuperscript{***} & 0.060\textsuperscript{***} \\
\midrule
\multirow{3}{*}{Acc (\%)}
  & Tree & 81.8 & 64.6 \\
  & Image & 55.3 & 49.6 \\
  & Fused & 83.5 & 66.4\textsuperscript{\textdagger} \\
  & Gain & +1.7 & +1.8 \\
\midrule
\multirow{2}{*}{Complement}
  & Score (\%) & 38.0 & 52.2 \\
  & F-Rescues & 9/52 & 4/19 \\
\bottomrule
\multicolumn{4}{l}{\footnotesize \textsuperscript{***}$p < 0.001$; \textsuperscript{\textdagger}Weighted addition for JM, concatenation for Neuron7} \\
\end{tabular}
\end{table}

\paragraph{Limitations}
While our complementarity analysis reveals that tree and image modalities capture distinct aspects of neuronal morphology (Pearson $< 0.09$, complementarity scores 38-52\%), simple fusion strategies show modest gains (+1.7-1.8\%). K-nearest neighbor evaluation in high-dimensional concatenated spaces suffers from the curse of dimensionality, where distance metrics become less discriminative. Additionally, uniform fusion strategies cannot adaptively weight modalities based on per-sample reliability, leading to cases where correct predictions from one modality are overridden by the weaker modality. The tree modality's superior performance (65-82\% vs 50-55\% for images) creates an imbalance where simple fusion struggles to selectively leverage the stronger modality. Detailed comparison of fusion strategies (concatenation, addition, weighted addition, PCA-based) and analysis of fusion rescue versus damage patterns are provided in the supplementary materials.

%% file: sec/5_ablation.tex
\section{Ablation Studies}

We conduct comprehensive ablation experiments to validate our design choices. We present key ablation studies here, with additional experiments provided in the supplementary material.

\noindent \textbf{Multi-channel encoding.} 
We evaluate the contribution of each channel in our persistence image representation (Table~\ref{tab:channel_ablation}). Single-channel configurations show that R (unweighted density) and G (persistence-weighted) achieve 60.7\% and 60.5\% respectively, outperforming B (radius-weighted) at 58.4\%. Two-channel combinations provide modest improvements, with RB reaching 61.7\%. The full RGB configuration substantially outperforms all partial combinations at 63.3\% average accuracy---a 2.6 percentage point gain over the best single channel.

\begin{table}[h]
\small
\centering
\caption{Ablation study on multi-channel persistence image encoding. R encodes unweighted density, G encodes persistence-weighted density, and B encodes radius-weighted density.}
\begin{tabular}{l|cccc}
\toprule
\textbf{Channel} & \textbf{ACT-4} & \textbf{BIL-6} & \textbf{JML-4} & \textbf{AVG} \\
\midrule
B only & 39.4 & 73.2 & 62.5 & 58.4 \\
G only & 45.6 & 73.6 & 62.2 & 60.5 \\
R only & 46.1 & 72.3 & 63.7 & 60.7 \\
GB & 44.7 & 74.5 & 62.5 & 60.6 \\
RB & 45.4 & 74.5 & 65.2 & 61.7 \\
RG & 46.1 & 71.9 & 62.5 & 60.2 \\
\midrule
\rowcolor{green!15} RGB & 47.9 & 75.3 & 66.7 & 63.3 \\
\bottomrule
\end{tabular}
\label{tab:channel_ablation}
\end{table}

\noindent \textbf{Augmentation strategies.}
Table~\ref{tab:augmentation_accuracy} reports ablation results on ACT-4 using DinoV2-small. The best accuracy (49.31\%) is achieved by combining all three persistence-based augmentations (Birth Death jitter, Persistence Scaling, Radius Perturbation), surpassing the no-augmentation baseline (40.97\%) by +8.34 points. Among individual methods, Radius Perturbation yields the highest gain (46.53\%), and the strongest two-augmentation variant is Persistence Scaling + Radius Perturbation (47.92\%).

\begin{table}[htbp]
\centering
\small
\caption{Augmentation Methods and Their Accuracy}
\label{tab:augmentation_accuracy}
\begin{tabular}{cccccc}
\toprule
\textbf{No Aug} & \textbf{BD jitter} & \textbf{P-Scale} & \textbf{R-Pert} & \textbf{ACT-4 (Acc \%)} \\
\midrule
\checkmark &            &            &            & 40.97 \\
           & \checkmark &            &            & 45.14 \\
           &            & \checkmark &            & 42.36 \\
           &           &             & \checkmark & 46.53 \\
           & \checkmark & \checkmark &            & 41.67 \\
          & \checkmark &            &    \checkmark & 43.06 \\
           &            & \checkmark & \checkmark & 47.92 \\
\rowcolor{green!15} & \checkmark & \checkmark & \checkmark & \textbf{49.31} \\
\bottomrule
\end{tabular}
\end{table}

%% file: sec/6_conclusion.tex
\section{Conclusion}
\label{sec:conclusions}
We presented \textbf{GraPHFormer}, a multimodal framework that unifies graph and topological representations of neuronal morphology via CLIP-style contrastive learning. By aligning a TreeLSTM-based graph encoder with a DINOv2-based persistence image encoder, the method captures complementary geometric and topological cues, achieving state-of-the-art results on most benchmarks. Our multi-channel persistence encoding and persistence-space augmentation proved key to robust cross-dataset generalization and self-supervised performance. We plan to extend GraPHFormer with cross-modal attention fusion and adaptive modality weighting to enhance per-sample reliability. Finally, incorporating 3D volumetric and EM-derived modalities and releasing an open multimodal benchmark will further advance reproducible computational connectomics~\cite{kanari2025mice}.

%% file: sec/acknowledgment.tex
\section{Acknowledgments} 
This publication was funded by the PPM7th Cycle grant (PPM 07-0409-240041, AMAL-For-Qatar) from the Qatar Research Development and Innovation Council (QRDI), a member of the Qatar Foundation. The findings herein reflect the work and are solely the responsibility, of the authors.

%% file: sec/X_suppl.tex
\clearpage
\setcounter{page}{1}
\maketitlesupplementary

\begin{abstract}
Quantitative analysis of neural morphology is central to understanding circuit development, computation, and pathology. Current methods often analyze topology or graph structure in isolation. We introduce GraPHFormer, a multimodal architecture that combines topological and graph representations through contrastive learning. The vision branch processes a three-channel persistence image encoding unweighted, persistence-weighted, and radius-weighted densities via DINOv2-ViT-S. In parallel, a TreeLSTM encoder captures geometric and radial attributes from skeleton graphs. Both project to a shared embedding space trained with symmetric InfoNCE loss. We evaluate GraPHFormer on six benchmarks (BIL-6, ACT-4, JML-4 , N7, M1-Cell, M1-REG) under self-supervised and supervised settings, demonstrating consistent improvements over topology-only, graph-only, and morphometrics baselines. We further demonstrate practical utility through cross-domain transfer between neuronal and glial morphologies and embedding space analysis.
\end{abstract}

\section{Overview}

This supplementary document provides additional experimental details, ablation studies, and analyses that complement the main paper. Specifically, we present:

\begin{itemize}
\item \textbf{Additional ablation studies} (Section~\ref{sec:supp_ablation}): We evaluate image encoder architectures, analyze the statistical independence of RGB channels through correlation analysis, and visualize individual channel contributions.

\item \textbf{Cross-domain generalizability} (Section~\ref{sec:cross_domain}): We assess transfer learning between neuronal and glial morphologies across species, demonstrating that the learned representations capture organizational principles that generalize across cell classes.

\item \textbf{Embedding space visualization} (Section~\ref{sec:embedding_viz}): We present t-SNE projections of learned embeddings across multiple datasets, revealing clustering patterns that reflect morphological organization.

\item \textbf{Cross-modal retrieval analysis} (Section~\ref{sec:retrieval}): We evaluate bidirectional retrieval between tree graphs and persistence images, characterizing the alignment and information asymmetry between modalities.

\item \textbf{Alternative training strategies} (Section~\ref{sec:moco}): We explore MoCo-style training with various fusion mechanisms, comparing attention-based and simple fusion strategies.

\item \textbf{Visual correspondence} (Section~\ref{sec:visual_corr}): We provide qualitative examples illustrating the relationship between tree structures and their persistence image representations.
\end{itemize}

These analyses provide deeper insights into architectural choices, learned representations, and cross-domain applicability.

\section{Additional Ablation Studies}
\label{sec:supp_ablation}

\subsection{Image Encoder Architecture}

Table~\ref{tab:ablation} compares six image encoder architectures for neuron morphology representation learning, using TreeLSTM as the tree encoder across all experiments. We evaluate DinoV2-S, ResNet18, ResNet50, SmallViT, and two hybrid variants (R18-ViT and R50-ViT) that replace the last two ResNet layers with vision transformer blocks. Models are trained for 50 epochs with self-supervised learning and evaluated every 5 epochs using frozen k-NN classification. Results are averaged over 5 random seeds. DinoV2-S achieves the best average performance (70.6\%), excelling on BIL (87.0\%) and JML-4 (74.3\%), while ResNet18 performs best on ACT (54.2\%).

\begin{table}[h]
\centering
\caption{Ablation study for image encoder selection across different architectures. Models are trained for 50 epochs using self-supervised learning and evaluated every 5 epochs with frozen k-NN classification on three benchmark datasets. Hybrid variants (R18-ViT, R50-ViT) replace the last two ResNet layers with ViT blocks. Results are averaged over 5 random seeds with best performance in \textbf{bold}.}
\label{tab:ablation}
\small
\begin{tabular}{l|ccc|c}
\toprule
\textbf{Model} & \textbf{ACT-4} & \textbf{BIL-6} & \textbf{JML-4} & \textbf{Average} \\
\midrule
R18-ViT & 49.7 ± 1.4 & 85.7 ± 0.0 & 72.6 ± 4.0 & 69.3 ± 1.8 \\
R50-ViT & 51.0 ± 1.4 & 84.9 ± 0.7 & 71.5 ± 0.7 & 69.1 ± 0.9 \\
ResNet50 & 52.9 ± 3.3 & 84.5 ± 1.8 & 71.5 ± 2.1 & 69.7 ± 2.4 \\
SmallViT & 49.6 ± 1.5 & 86.0 ± 1.1 & 70.3 ± 2.5 & 68.6 ± 1.7 \\
ResNet18 & \textbf{54.2 ± 1.0} & 84.9 ± 1.2 & 71.7 ± 1.7 & 70.3 ± 1.3 \\
\rowcolor{green!15} DinoV2-S & 50.6 ± 0.9 & \textbf{87.0 ± 2.1} & \textbf{74.3 ± 1.9} & \textbf{70.6 ± 1.6} \\
\bottomrule
\end{tabular}
\end{table}

\subsection{RGB Channel Independence Analysis}

We validate our multi-channel encoding by extracting features from single-channel persistence images and computing pairwise Spearman correlations on BIL-6. Table~\ref{tab:channel_correlation} shows moderate correlations (0.35--0.45, average 0.41). The G--B pair (persistence versus radius) exhibits lowest correlation (0.354), demonstrating that branch length and thickness encode complementary geometry. While not perfectly orthogonal ($\rho < 0.2$), moderate correlations are expected since all channels derive from the same underlying morphology. The ablation results presented in the main paper show that RGB (63.3\%) substantially outperforms single ($\leq$60.7\%) or dual ($\leq$61.7\%) channel configurations, confirming that all three channels contribute unique information despite moderate correlation.

\begin{table}[h]
\centering
\caption{Pairwise correlation between RGB channel features (BIL-6 dataset). Moderate correlations indicate related but distinct morphological aspects. The G--B pair (0.354) is most orthogonal, showing that persistence and radius capture complementary information.}
\label{tab:channel_correlation}
\small
\begin{tabular}{lccc}
\toprule
\textbf{Channel Pair} & \textbf{Correlation} & \textbf{Interpretation} \\
\midrule
R -- G & 0.433 & Moderate \\
R -- B & 0.447 & Moderate \\
G -- B & 0.354 & Lower (more orthogonal) \\
\midrule
\rowcolor{green!15} \textbf{Average} & \textbf{0.411} & \textbf{Moderate independence} \\
\bottomrule
\end{tabular}
\end{table}

\begin{figure}[h]
    \centering
    \includegraphics[width=0.98\linewidth]{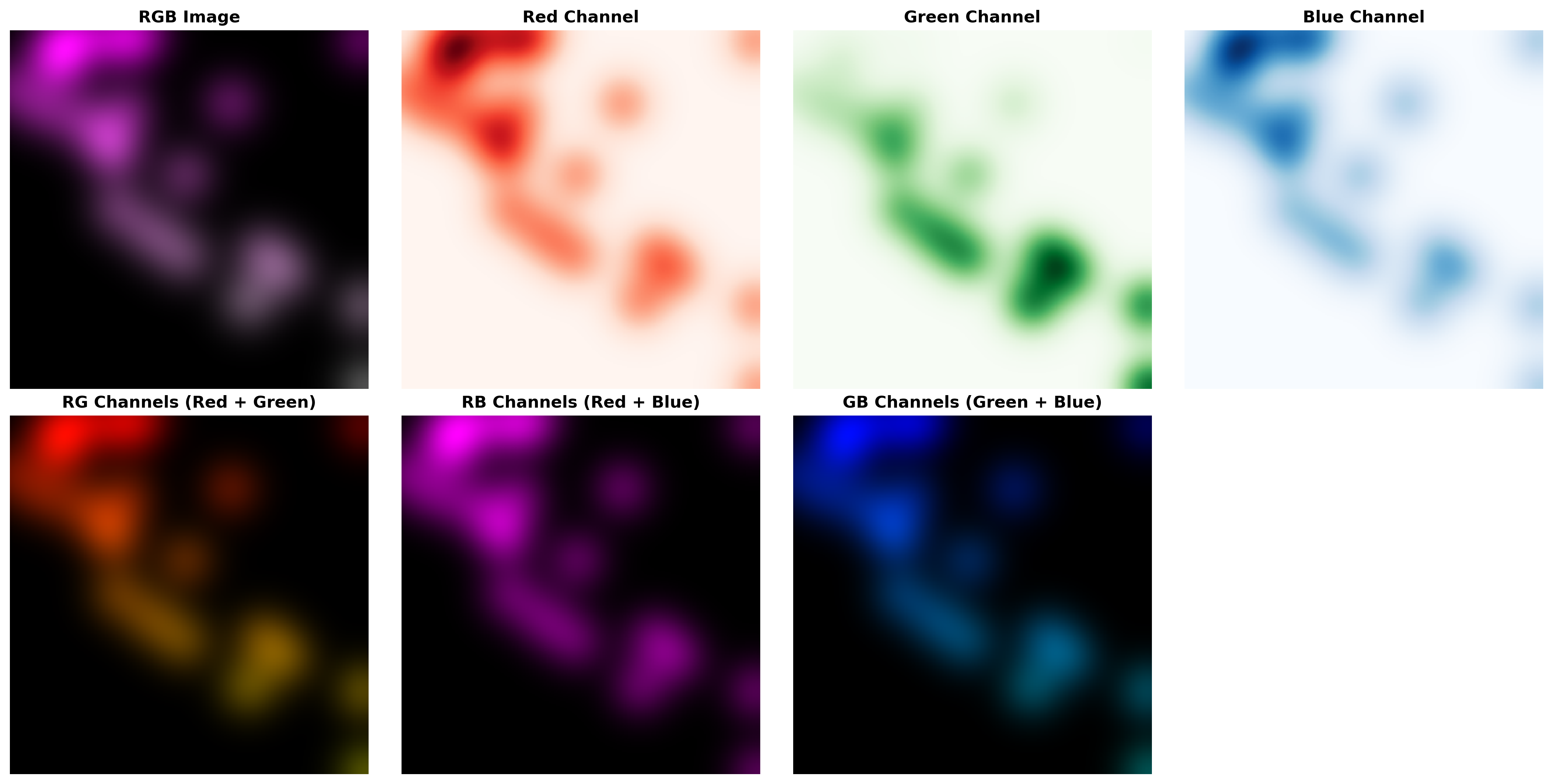}
    \caption{Visualization of RGB persistence image encoding and channel ablations. Each channel encodes distinct topological features. The R channel exhibits high intensity in the upper region while the G channel shows high intensity in the middle area. Channel combinations (RG, RB, GB) demonstrate how different pairings capture complementary information, with the full RGB representation providing comprehensive encoding of topological features.}
    \label{fig:channels_encoding}
\end{figure}

Figure~\ref{fig:channels_encoding} visualizes the individual and combined RGB channel encodings. Each channel encodes distinct topological features: the R channel shows high intensity in regions corresponding to dense branching near the soma, while the G channel emphasizes persistent branches in intermediate regions. Channel combinations (RG, RB, GB) demonstrate how different pairings capture complementary information, with the full RGB representation providing comprehensive encoding of topological features.

\begin{figure}
    \centering
    \includegraphics[width=0.49\linewidth]{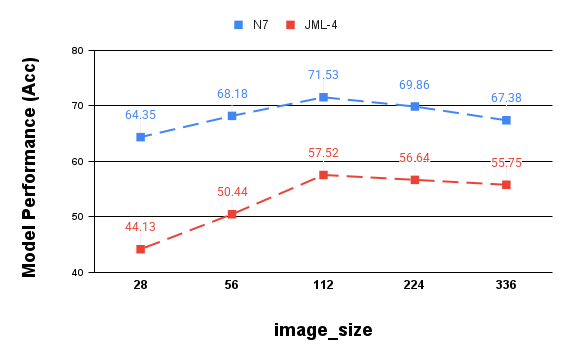}
    \includegraphics[width=0.49\linewidth]{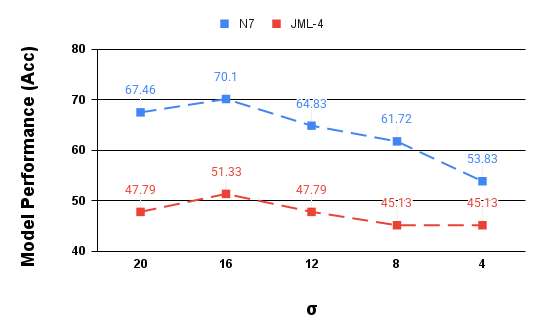}
    \caption{Ablation study on Image size (left) and Gaussian Kernel (right) using Image Encoder Only.}
    \label{fig:image_kernel}
\end{figure}

\subsection{Image resolution and Gaussian kernels.}
We evaluated the impact of image resolution and Gaussian kernel size $\sigma$ on model performance using supervised learning on the Neuron7 and JML datasets. Image resolutions ranged from 28 to 336 pixels (in multiples of 14 to accommodate the DinoV2 encoder), while kernel values $\sigma$ ranged from 4 to 20. Results are presented in Figure~\ref{fig:image_kernel}. Performance peaked at resolution 112 for both datasets, achieving 71.53\% on Neuron7 and 57.52\% on JML-4, then declined at higher resolutions. For Gaussian kernels, optimal performance occurred at $\sigma=16$ with 70.1\% on Neuron7. Based on these results, we use an image resolution of $112 \times 112 \times 3$ and $\sigma=16$ for all subsequent experiments.
\subsection{Embedding, Batch size and Projection Head}
\noindent\textbf{Embedding Size.} We ablate the projection embedding 
dimension across $\{32, 64, 128, 256\}$, trained jointly on BIL-6, ACT-4, 
and JML-4 for 50 epochs under the self-supervised scheme. EM-128 achieves 
the best overall performance, with consistent gains over smaller dimensions, 
while EM-256 slightly degrades on ACT-4, suggesting that overly large 
embedding spaces may hurt generalization on smaller datasets.
\begin{table}[h]
\centering
\caption{Ablation on embedding size during self-supervised training. }
\label{tab:ablation_emb}
\begin{tabular}{l|cccc}
\hline
Data & EM-32 & EM-64 & EM-128 & EM-256 \\
\hline
ACT-4 & 53.47 & \textbf{56.25} & \textbf{56.25} & 51.39 \\
BIL-6 & 84.42 & 85.71 & \textbf{86.20} & 85.71 \\
JML-4 & 65.49 & 65.49 & 67.26 & \textbf{69.03} \\
\hline
\end{tabular}
\end{table}

\noindent\textbf{Batch Size.} We evaluate batch sizes $\{64, 128, 256\}$ 
under the same joint training protocol. BS-128 yields the best results 
across all three datasets, as larger batches provide more within-batch 
negatives for contrastive learning, while BS-256 shows slight degradation, 
likely due to reduced gradient diversity at our dataset scale.

\begin{table}[h]
\centering
\caption{Ablation on batch size during self-supervised training.}
\label{tab:ablation_bs}
\begin{tabular}{l|ccc}
\hline
Data & BS-64 & BS-128 & BS-256 \\
\hline
ACT-4 & 53.47 & \textbf{56.25} & 54.17 \\
BIL-6 & 83.12 & \textbf{86.20} & 83.12 \\
JML-4 & 64.60 & 67.26 & \textbf{68.33} \\
\hline
\end{tabular}
\end{table} 
\noindent\textbf{Projection Head.} We compare an MLP projection head against 
a single linear layer. The MLP consistently outperforms the linear layer 
across all datasets, with the most notable gap on ACT-4 (56.25\% vs.\ 
52.08\%), confirming that non-linear projections better capture the complex 
morphological feature interactions needed for effective contrastive 
representation learning.
\begin{table}[h]
\centering
\caption{Ablation on projection head architecture: MLP vs.\ single linear layer.}
\label{tab:ablation_head}
\begin{tabular}{l|cc}
\hline
Data & MLP & One Layer \\
\hline
ACT-4 & \textbf{56.25} & 52.08 \\
BIL-6 & \textbf{86.20 }& 84.42 \\
JML-4 & \textbf{67.26} & 63.72 \\
\hline
\end{tabular}
\end{table}

\section{Cross-Domain Generalizability: Neuron-to-Glia Transfer}
\label{sec:cross_domain}

To assess whether GraPHFormer learns representations that generalize across cell classes, we conducted cross-domain transfer experiments between neuronal and glial morphologies. We obtained glia reconstructions from NeuroMorpho.Org~\cite{ascoli2007neuromorpho} spanning four species: mouse (7,000 samples), rat (2,149), semipalmated sandpiper (1,784), and semipalmated plover (992). Our evaluation protocol consisted of two transfer scenarios: (1) training on neuronal morphologies and testing on glia for species classification, and (2) training on glia and evaluating across six established neuronal benchmarks (BIL-4, JML-4 , ACT-4, N7, M1-Cell, M1-REG). All models were trained for 50 epochs using self-supervised learning and evaluated with frozen k-NN classification following the TreeMoCo protocol.

\begin{table}[h]
\centering
\small
\caption{Cross-domain transfer performance between neuronal and glial morphologies. \textbf{Left column}: Model trained on neuron data, tested on glia (species classification). \textbf{Right columns}: Model trained on glia, tested on neuron datasets (cell-type classification). Results demonstrate transfer capability despite morphological differences between cell classes.}
\label{tab:generalizability}
\begin{tabular}{lcc}
\hline
\textbf{Test Dataset} & \textbf{Train: Neuron} & \textbf{Train: Glia} \\
\hline
Glia (species) & 78.87 & 86.94 \\
\midrule
BIL-4 & - & 81.82 \\
JML-4  & - & 68.14 \\
ACT-4 & - & 46.52 \\
N7 & - & 82.78 \\
M1-REG & - & 71.08 \\
M1-Cell & - & 72.84 \\
\hline
\end{tabular}
\end{table}

Table~\ref{tab:generalizability} reveals cross-domain transfer despite morphological differences between neurons and glia. When trained exclusively on neuronal data, GraPHFormer achieves 78.87\% accuracy on glia species classification—8 percentage points below the glia-trained model (86.94\%). This demonstrates that multimodal topological-structural features transfer across different cell classes.

The reverse transfer—glia-to-neuron—yields competitive performance across several neuronal benchmarks. On N7 (82.78\%) and BIL-4 (81.82\%), the glia-trained model approaches within-domain performance, suggesting that branching topology and radial geometry encode organizational principles that are shared across cell classes. Performance on JML-4  (68.14\%) and cortical datasets (M1-REG: 71.08\%, M1-Cell: 72.84\%) remains competitive, though the lower ACT-4 result (46.52\%) suggests that cortical layer discrimination may benefit more from neuron-specific features.

These findings indicate two properties of GraPHFormer's learned representations: (1) the multimodal fusion of persistence images and graph structure captures morphological signatures that generalize across cell types, and (2) self-supervised contrastive learning discovers features that maintain utility across domain shifts. The ability to transfer between neurons and glia—morphologically distinct yet topologically related—suggests potential for few-shot learning scenarios and cross-species comparative studies where labeled data is limited.

\section{Embedding Space Visualization}
\label{sec:embedding_viz}

\begin{figure*}[h]
    \centering
    \includegraphics[width=0.48\linewidth]{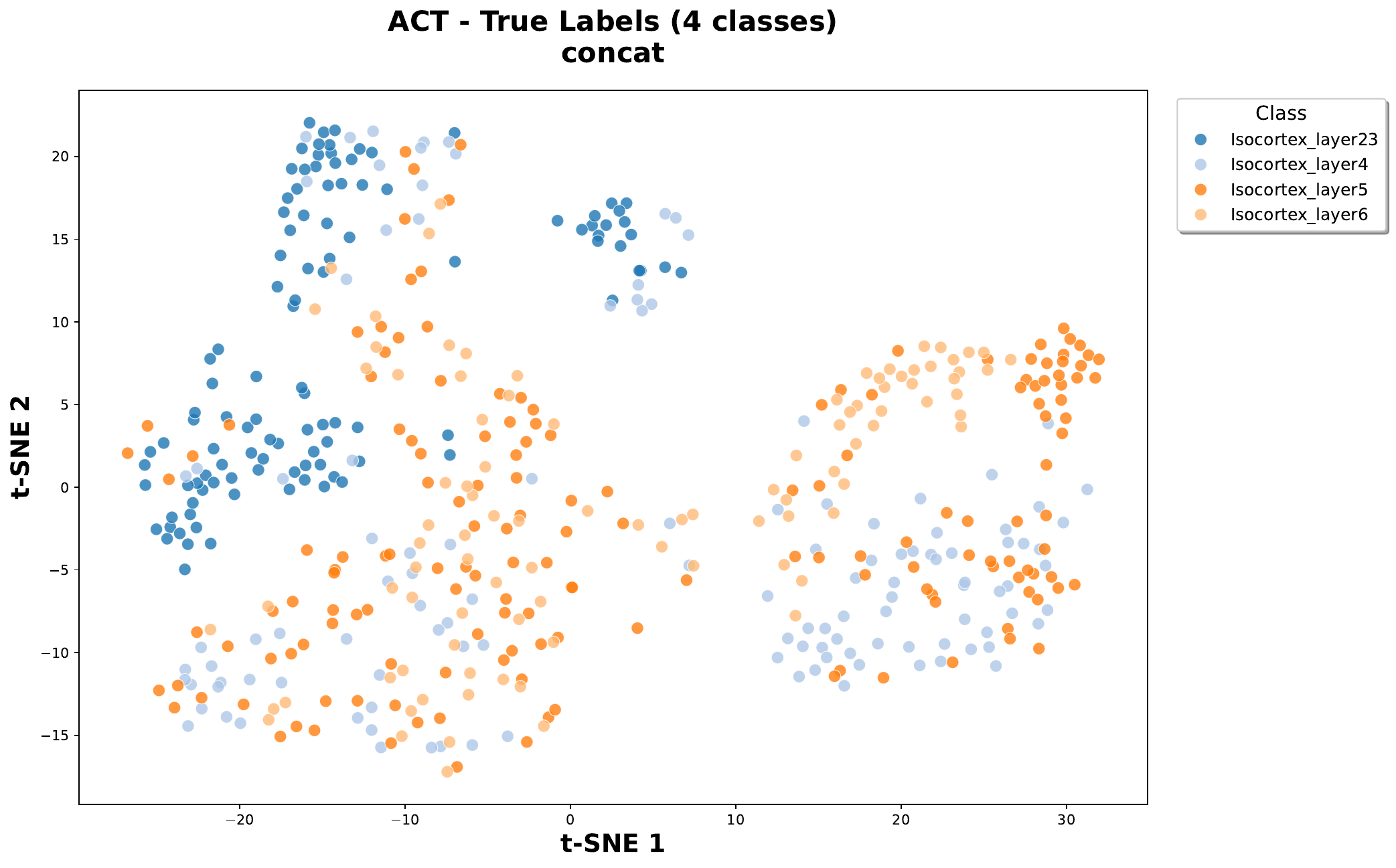}
    \includegraphics[width=0.48\linewidth]{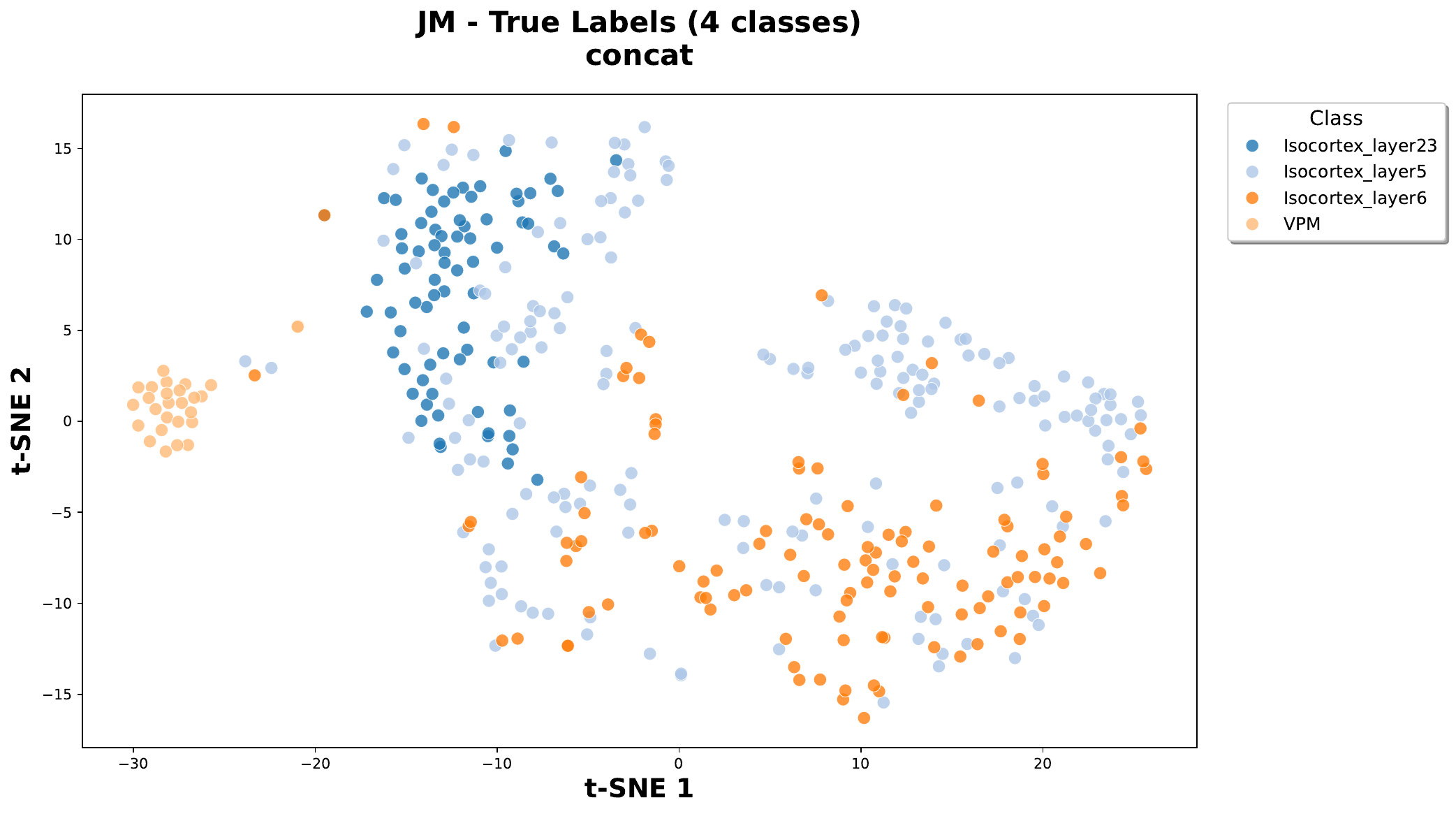}
    \includegraphics[width=0.48\linewidth]{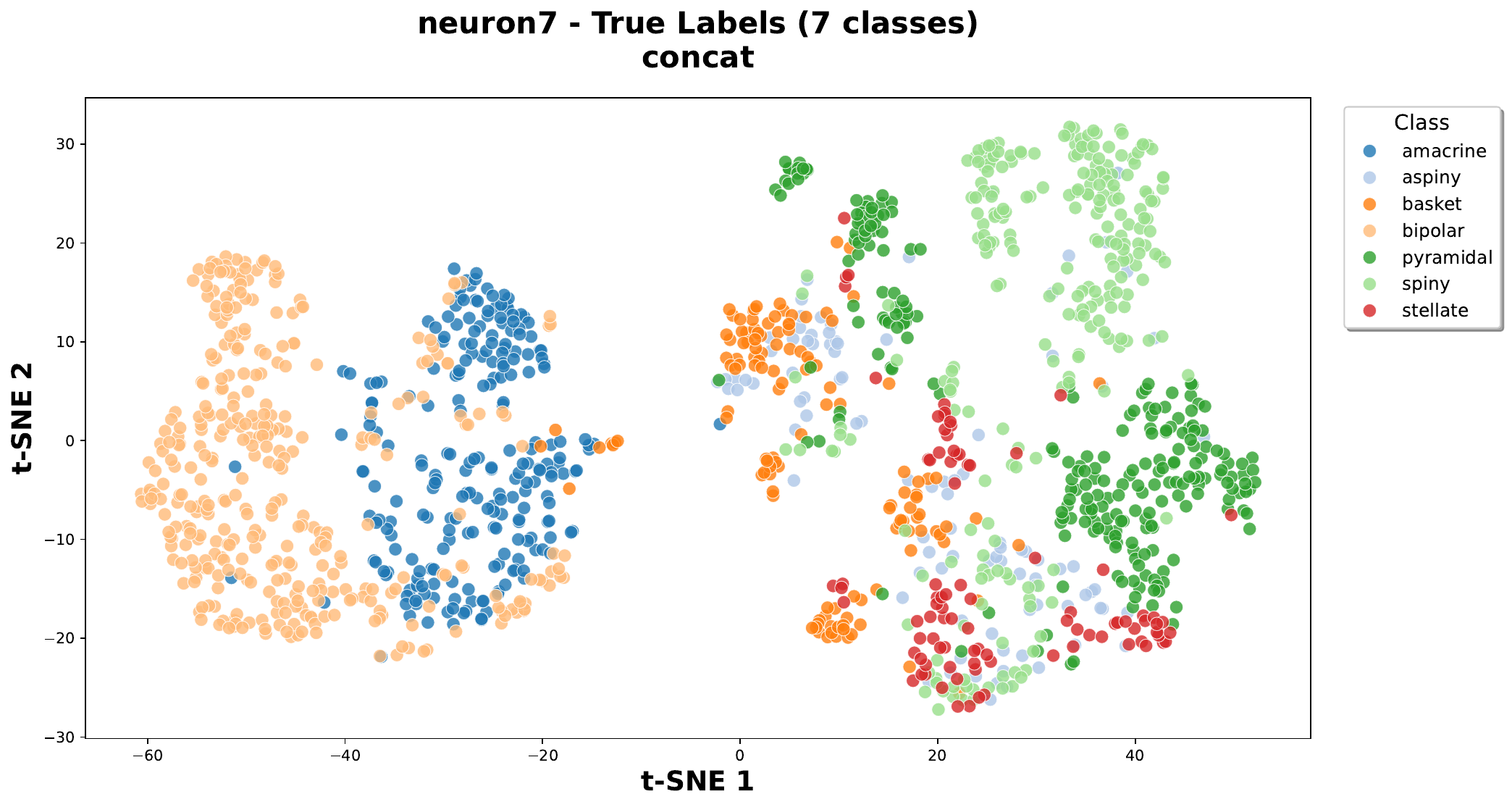}
    \includegraphics[width=0.48\linewidth]{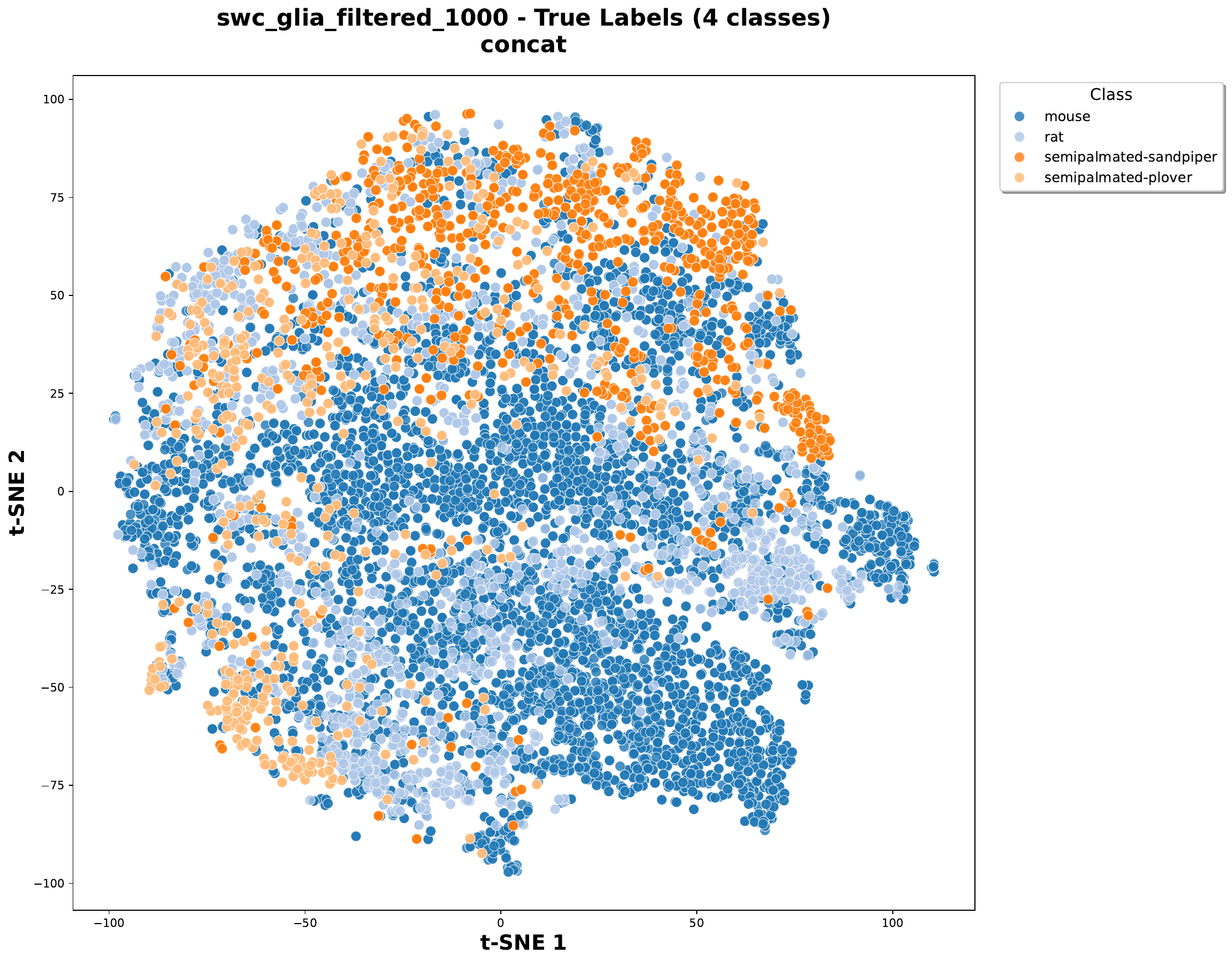}
    \caption{t-SNE visualization of GraPHFormer embedding spaces across four datasets. \textbf{Top left}: ACT-4 (cortical layers 2/3, 4, 5, 6) shows partial overlap reflecting morphological gradients between adjacent layers. \textbf{Top right}: JML-4 displays separation between thalamic VPM neurons (left cluster) and cortical projection neurons (center-right). \textbf{Bottom left}: Neuron7 exhibits distinct clusters for seven morphologically diverse cell types (bipolar, amacrine, basket, pyramidal, spiny, stellate), demonstrating discrimination of fundamental neuronal classes. \textbf{Bottom right}: Glia species classification (mouse, rat, semipalmated sandpiper/plover) shows overlapping distributions reflecting conserved morphological features across species. Embeddings are obtained by concatenating tree encoder and image encoder outputs after self-supervised contrastive pretraining. Clustering quality correlates with biological distinctiveness and classification accuracy.}
    \label{fig:tsne_clusters}
\end{figure*}

To qualitatively assess the learned representations, we visualized the GraPHFormer embedding space using t-SNE dimensionality reduction. Figure~\ref{fig:tsne_clusters} displays the concatenated multimodal embeddings (tree + image encoders) for four representative datasets: ACT-4 (cortical layers), JML-4 (cortical layers and thalamic neurons), Neuron7 (diverse cell types), and Glia (cross-species). The visualizations reveal clustering patterns that reflect the morphological organization captured by our framework.

For the \textbf{ACT-4 dataset} (cortical layer classification), neurons from layers 2/3, 4, 5, and 6 form partially overlapping clusters with some intermixing, particularly between layers 5 and 6. This overlap aligns with the challenging nature of layer-based classification (59.1\% accuracy in self-supervised setting) and reflects gradual morphological changes across adjacent cortical layers rather than discrete boundaries.

The \textbf{JML-4 dataset} exhibits clearer separation, with VPM (ventral posteromedial thalamic) neurons forming a distinct cluster on the left, while cortical neurons (layers 2/3, 5, 6) show moderate overlap in the center and right regions. This separation pattern is consistent with the higher classification accuracy (72.7\%) and demonstrates that GraPHFormer distinguishes between thalamic and cortical projection patterns, which exhibit more pronounced morphological differences than intra-cortical variations.

\textbf{Neuron7} shows well-organized clustering, with seven cell types forming separated, cohesive groups. Bipolar and amacrine cells (left), basket cells (center), and pyramidal/spiny neurons (right) occupy distinct regions of the embedding space. This organization (83.8\% accuracy, ±0.6\% variance) indicates that multimodal fusion captures morphological signatures that distinguish functionally diverse neuronal classes.

The \textbf{Glia dataset} (species classification) presents substantial overlap among mouse, rat, and two bird species (semipalmated sandpiper and plover). Despite this overlap—reflecting conserved glial morphologies across species—GraPHFormer achieves 86.94\% accuracy, indicating that the learned representations capture species-specific variations in glial process organization that are not immediately apparent in the 2D projection.

These visualizations demonstrate that GraPHFormer learns embedding spaces where morphologically similar cells cluster together, with separation quality correlating with both biological distinctiveness and quantitative classification performance.
\begin{table}[th]
\centering
\small
\caption{Ablation study of fusion strategies within the TreeMoCo framework for neuron morphological analysis. All variants employ ResNet-18 as the image encoder and TreeLSTM as the tree-structured encoder, jointly trained on the ACT-4, JML-4, and BIL-6 datasets using contrastive learning. Performance is evaluated on held-out test sets via frozen k-NN classification. We report average classification accuracy (in \%) ± standard deviation across three independent random seeds for each fusion method.}
\label{tab:fusion_ablation}
\begin{tabular}{lccc}
\toprule
\textbf{Fusion Strategy} & \textbf{ACT-4} & \textbf{BIL-6} & \textbf{JML-4} \\
\midrule
Bi-Attention & 54.39 $\pm$ 3.6 & 87.04 $\pm$ 0.7 & 73.59 $\pm$ 4.5 \\
Addition & 57.54 $\pm$ 4.2 & 84.5 $\pm$ 1.1 & 72.73 $\pm$ 1.3 \\
CAMME~\cite{naseem2025camme} & 52.28 $\pm$ 1.6 & 84.57 $\pm$ 2.1 & 71.86 $\pm$ 1.9 \\
Concatenation & 57.89 $\pm$ 1.1 & 83.33 $\pm$ 1.8 & 71.86 $\pm$ 1.9 \\
Gated & 58.25 $\pm$ 4.26 & 85.19 $\pm$ 0.7 & 72.30 $\pm$ 1.9 \\
MHCA & 58.90 $\pm$ 1.82 & 84.57 $\pm$ 1.07 & 67.10 $\pm$ 0.7 \\
\bottomrule
\end{tabular}
\end{table}

\section{Cross-Modal Retrieval Analysis}
\label{sec:retrieval}

\begin{figure*}[h]
    \centering
    \includegraphics[width=0.32\linewidth]{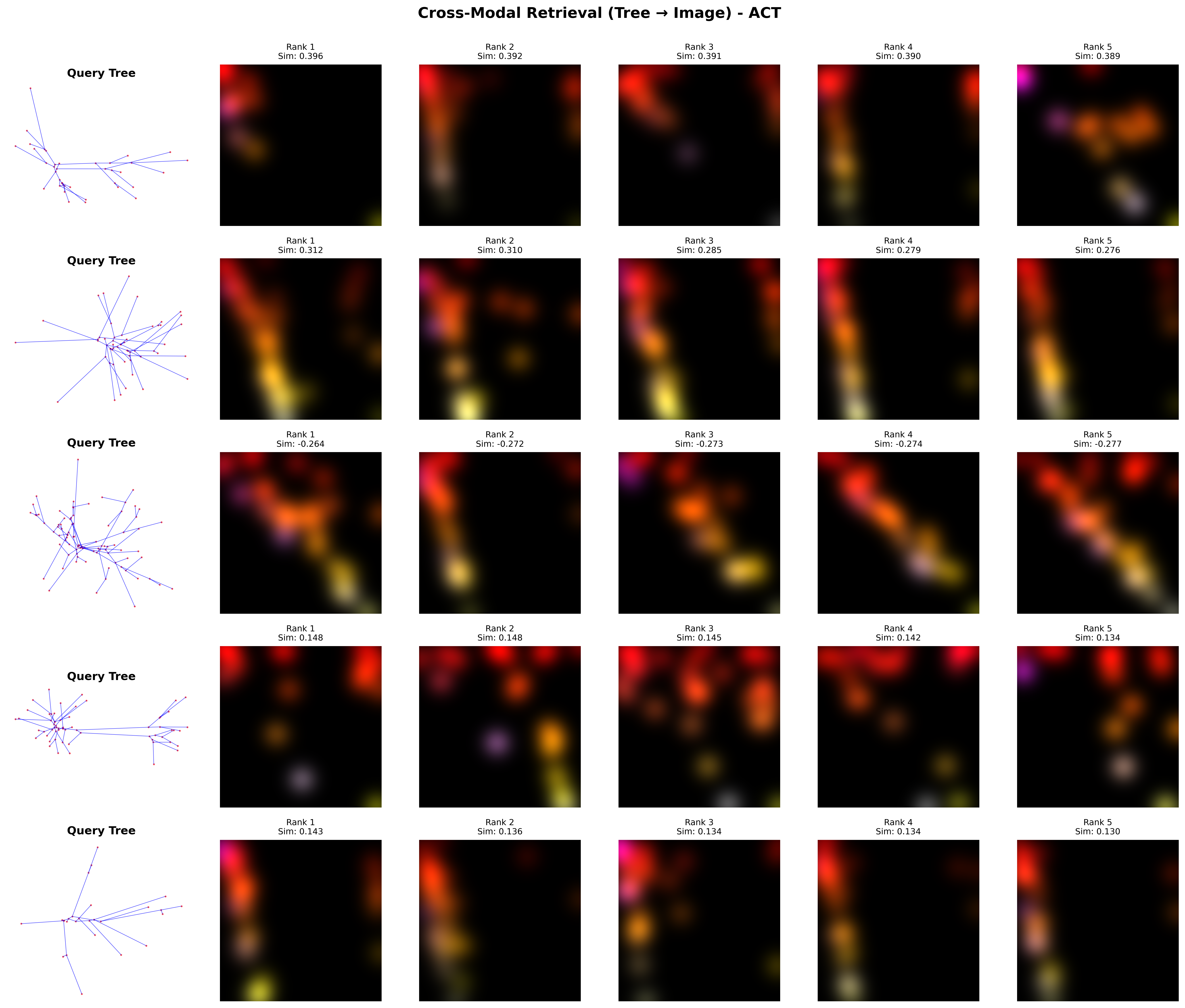}
    \includegraphics[width=0.32\linewidth]{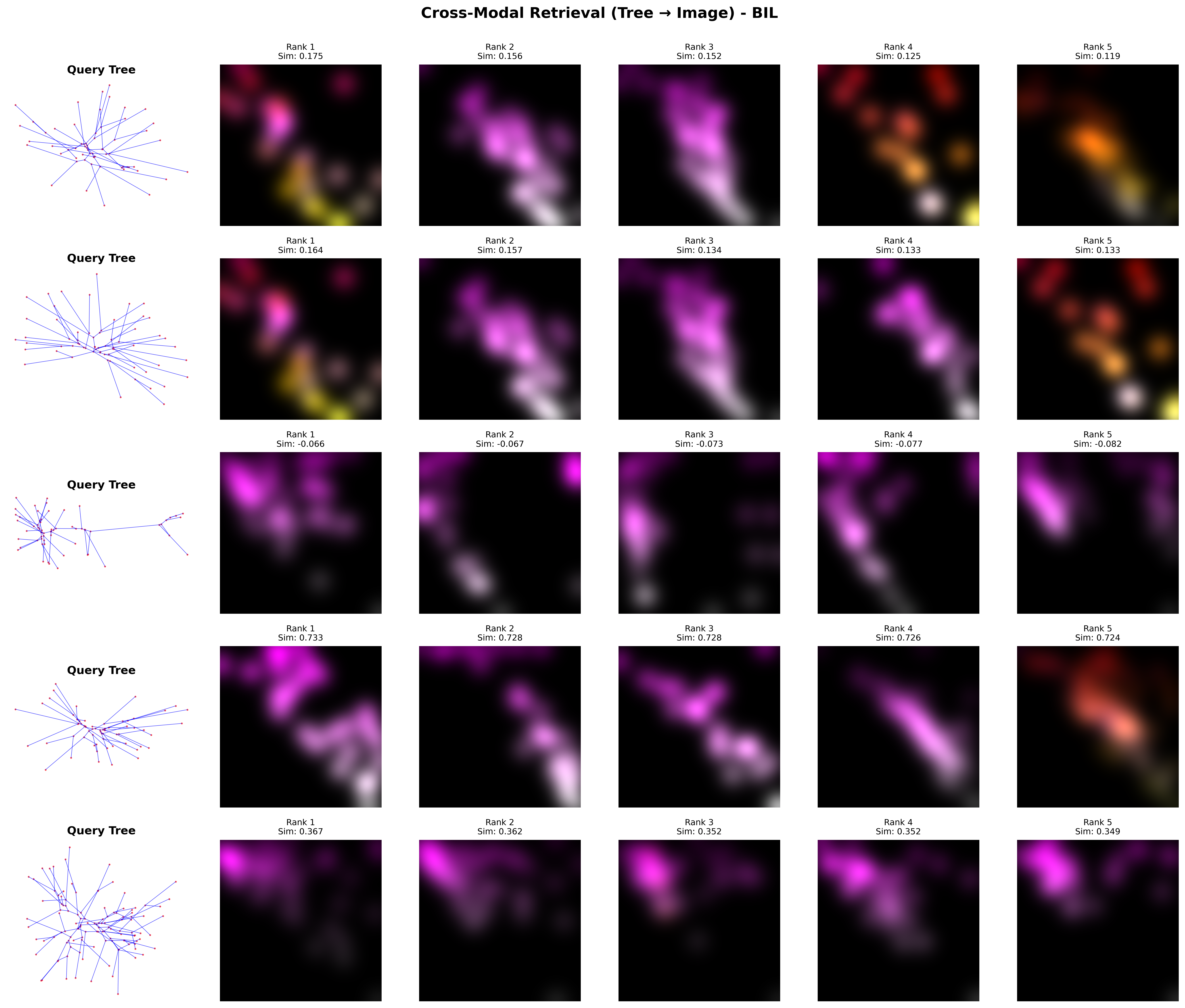}
    \includegraphics[width=0.32\linewidth]{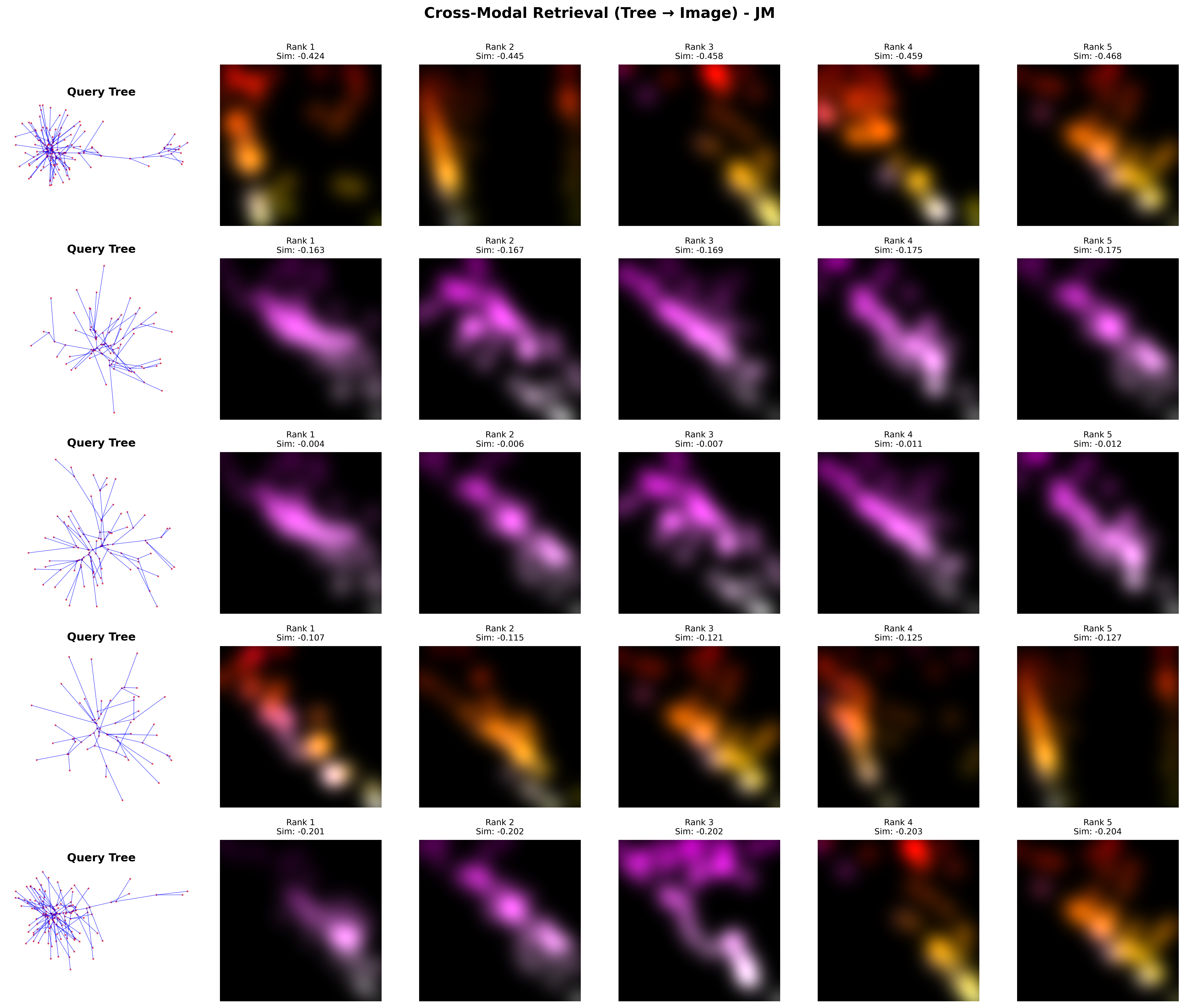}
    \\
    \vspace{0.2cm}
    \includegraphics[width=0.32\linewidth]{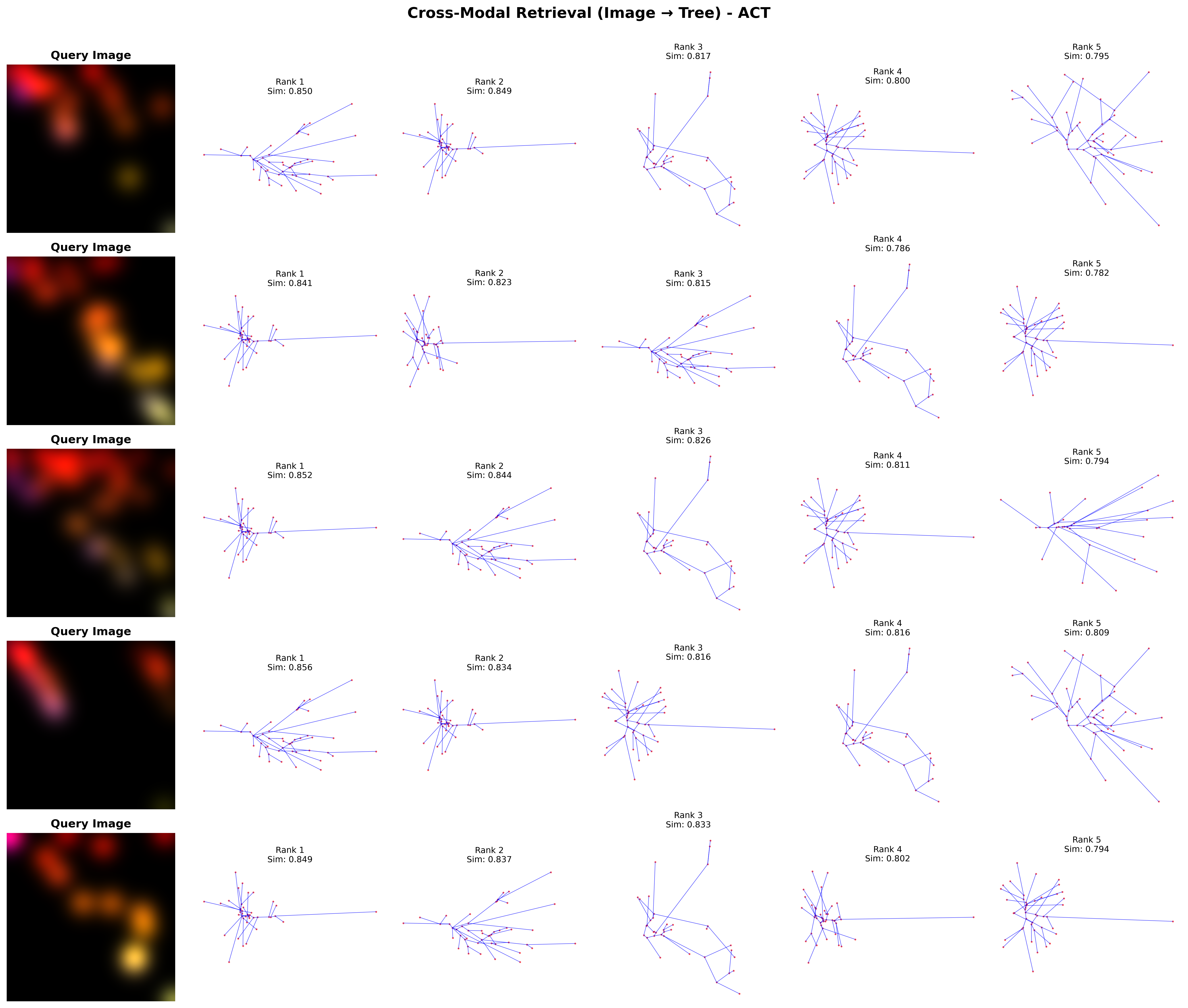}
    \includegraphics[width=0.32\linewidth]{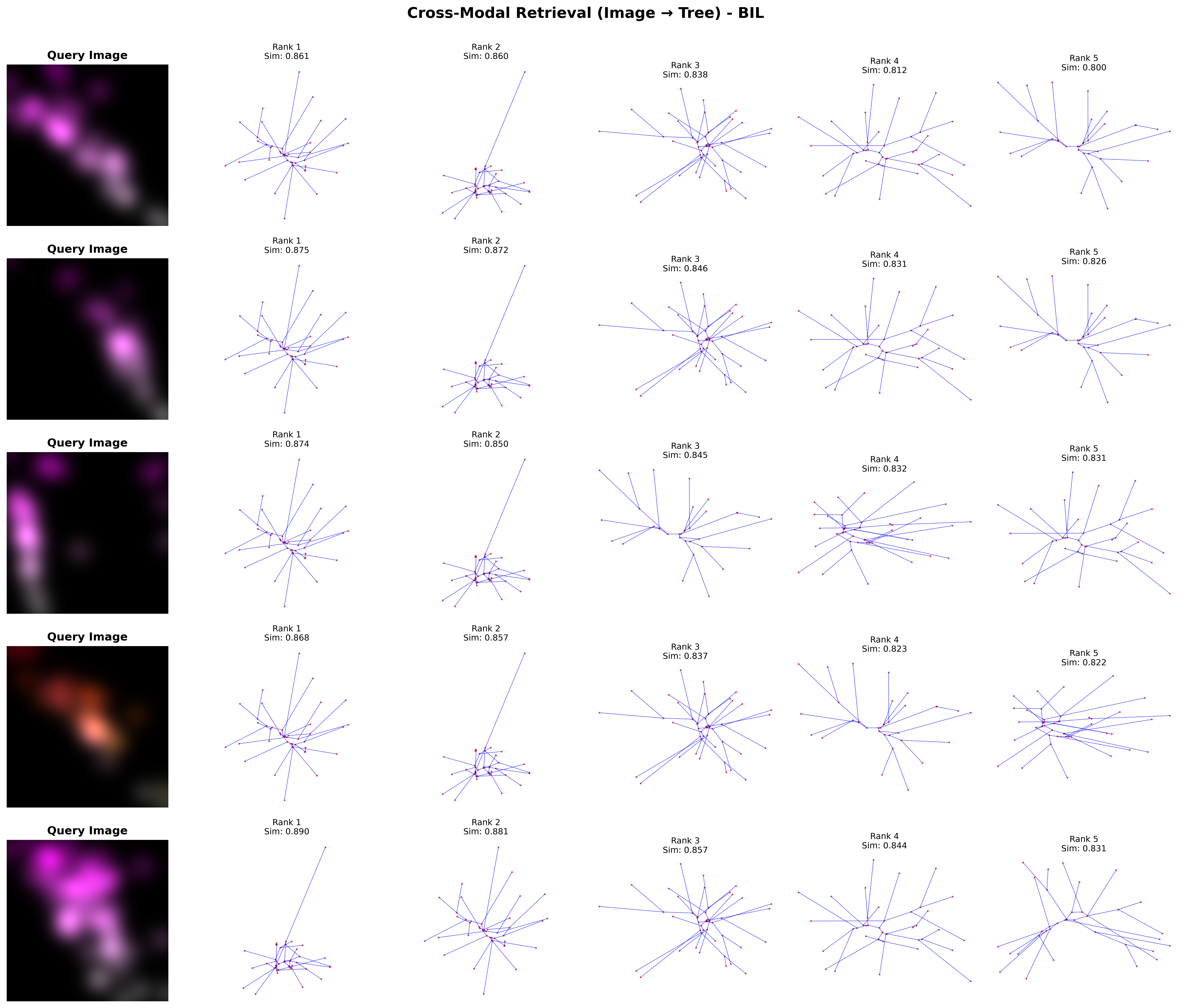}
    \includegraphics[width=0.32\linewidth]{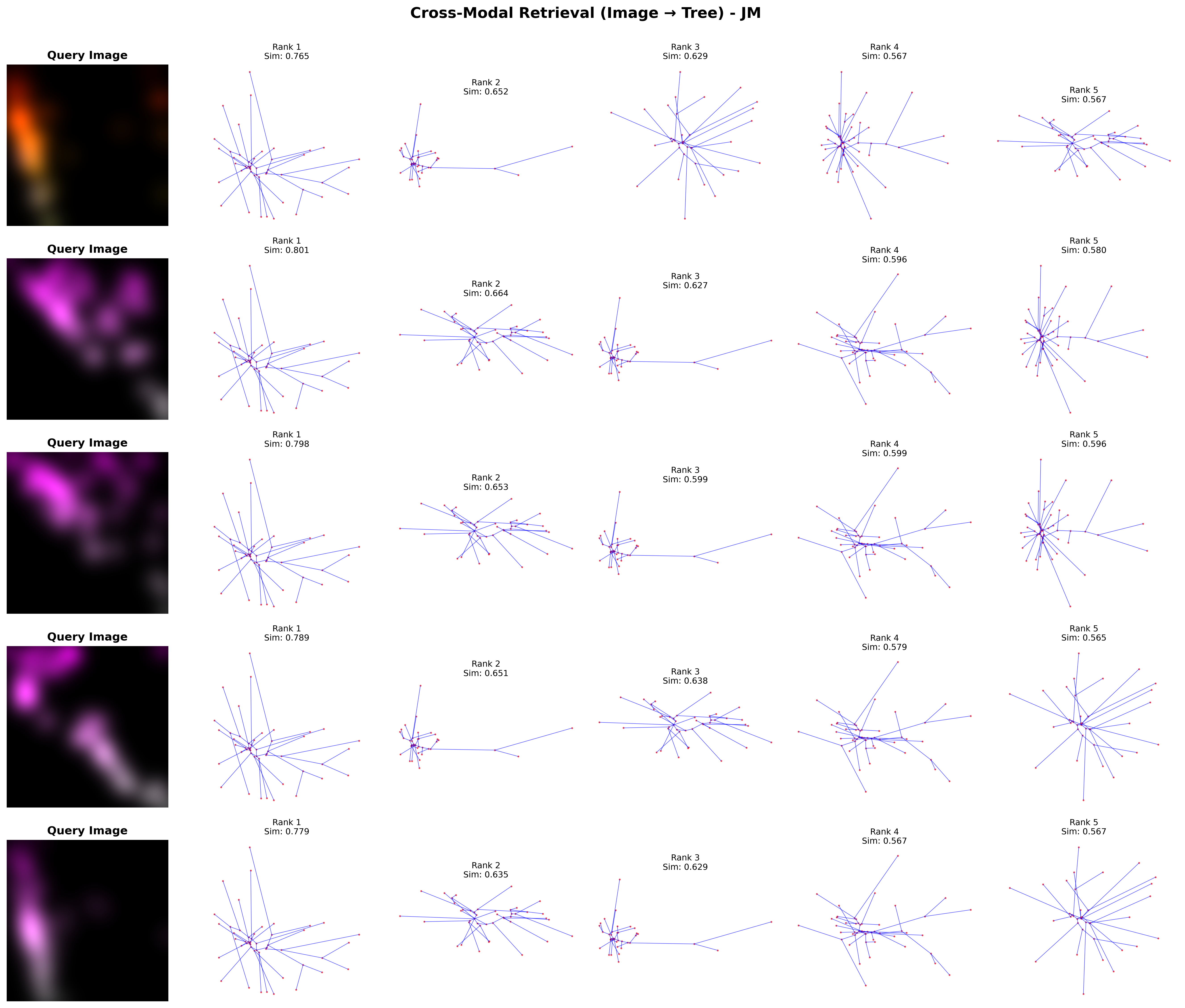}
    \caption{Bidirectional cross-modal retrieval demonstrating alignment between tree graphs and persistence images. \textbf{Top row}: Tree-to-image retrieval for ACT, BIL, and JML-4 datasets showing query tree graphs (left) and top-5 retrieved persistence images with cosine similarity scores. \textbf{Bottom row}: Image-to-tree retrieval showing query persistence images (left) and top-5 retrieved tree graphs. Note the asymmetry in similarity scores: image-to-tree retrieval achieves higher similarities (0.80-0.87) compared to tree-to-image retrieval (0.10-0.46), reflecting differences in information density between the graph and compressed topological representations.}
    \label{fig:cross_modal_retrieval}
\end{figure*}

To evaluate the alignment between tree and image representations, we performed bidirectional cross-modal retrieval experiments. For each query from one modality, we retrieve the top-5 nearest neighbors from the other modality using cosine similarity in the learned embedding space. Figure~\ref{fig:cross_modal_retrieval} shows retrieval results for ACT, BIL, and JML-4 datasets in both directions.

\paragraph{Asymmetric similarity scores.} Image-to-tree retrieval consistently achieves higher cosine similarities (0.80-0.90 across datasets) compared to tree-to-image retrieval (BIL: 0.12-0.73, JM: 0.00-0.46, ACT: 0.13-0.39). This asymmetry stems from differences in information content: tree graphs preserve detailed geometric information (coordinates, radii, exact connectivity), while persistence images compress morphology into topological summaries through filtration and Gaussian smoothing.

\paragraph{Information bottleneck.} When querying with a tree, the detailed geometric specification may not find exact matches in the compressed persistence image space, resulting in lower similarities. Conversely, when querying with a persistence image, multiple geometrically distinct trees sharing similar topological features can match the query pattern, yielding higher similarities.

\paragraph{Semantic alignment preserved.} Despite lower absolute scores in tree-to-image retrieval, retrieved neighbors generally belong to the correct morphological class, as evidenced by consistent patterns in persistence images and similar tree structures. This indicates successful alignment through contrastive learning, even with modalities of different information density.

These results demonstrate that both modalities contribute complementary information—graphs provide geometric precision while images offer topological invariance—supporting the effectiveness of our multimodal fusion approach.

\section{MoCo-Style Training}
\label{sec:moco}

We adopted the TreeMoCo framework~\cite{chen:2022:treemoco} to evaluate MoCo-style training of our multimodal approach. We integrated the tree encoder and image encoder through various fusion strategies, including multi-headed cross-attention (MHCA), bi-directional cross-attention, addition, concatenation, gated fusion, and CAMME~\cite{naseem2025camme}. Our experimental setup and loss function closely follow TreeMoCo, where the objective is to maximize similarity between positive pairs while minimizing similarity between negative pairs.

We employed ResNet-18 as the image encoder and TreeLSTM as the tree encoder. Following TreeMoCo, we jointly trained the models on the BIL-6, ACT-4, and JML-4 datasets in a self-supervised learning scheme. Model performance was evaluated using frozen k-NN classification, with $k=20$ for BIL-6 and ACT-4, and $k=5$ for JML-4, consistent with the TreeLSTM evaluation protocol~\cite{chen:2022:treemoco}. All models were trained for 50 epochs using SGD with an initial learning rate of 0.06 and batch size 128.

Table~\ref{tab:fusion_ablation} presents the fusion strategy ablation results. Performance varies across datasets, with no single strategy dominating universally. On BIL-6, bi-directional attention achieves 87.04\% (±0.7), while MHCA performs best on ACT-4 (58.90\% ±1.82). Bi-directional attention performs well on JML-4 (73.59\% ±4.5), though with notably higher variance. Simple strategies like addition and gated fusion provide competitive results across all datasets, with addition achieving 57.54\% on ACT-4 and gated fusion reaching 58.25\%. These results indicate that fusion strategy selection depends on specific task characteristics, with attention-based mechanisms generally showing competitive performance but with dataset-dependent effectiveness.

\section{Visual Correspondence Between Modalities}
\label{sec:visual_corr}

\begin{figure*}[t]
    \centering
    \includegraphics[width=0.32\linewidth]{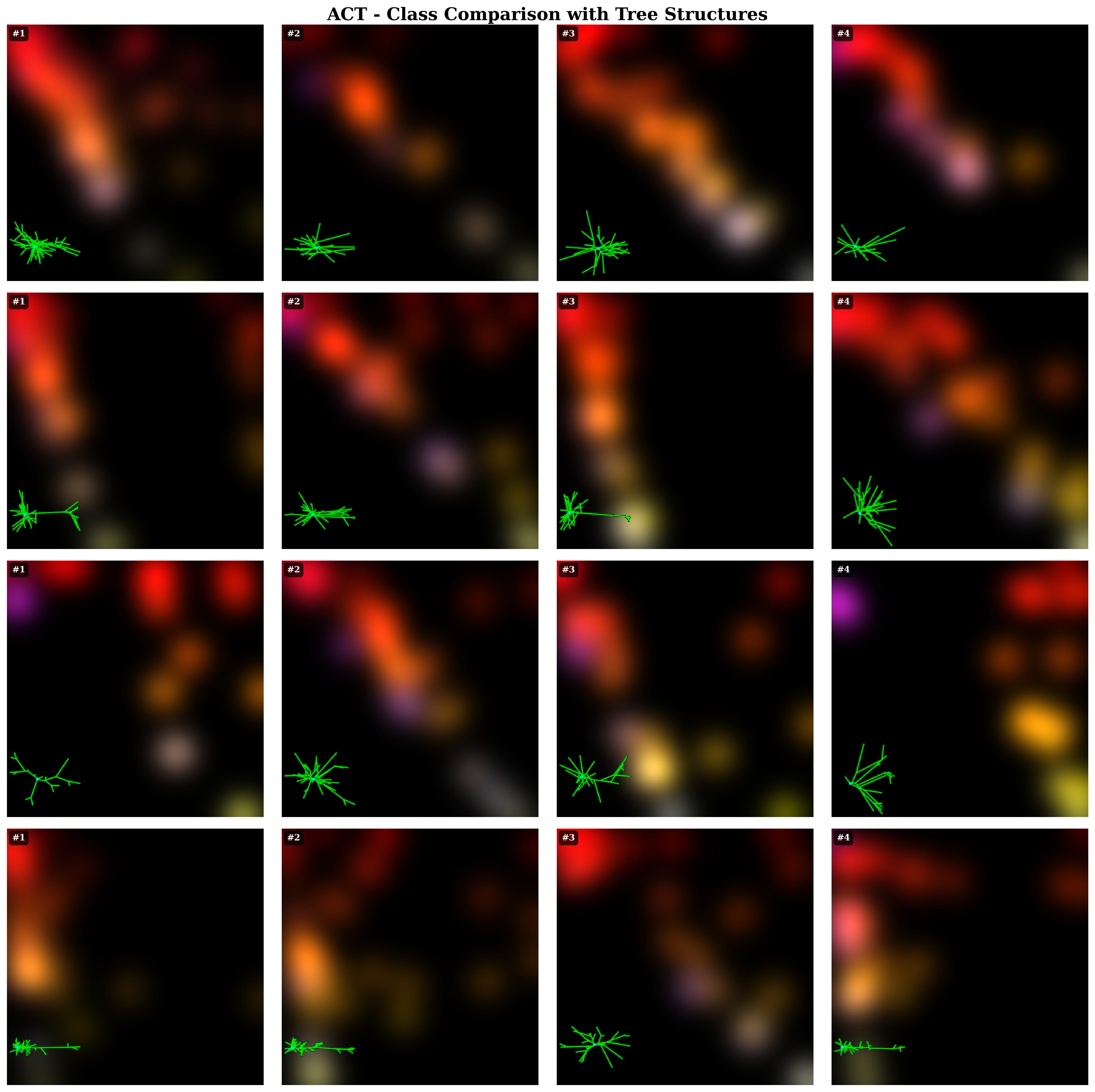}
    \includegraphics[width=0.32\linewidth]{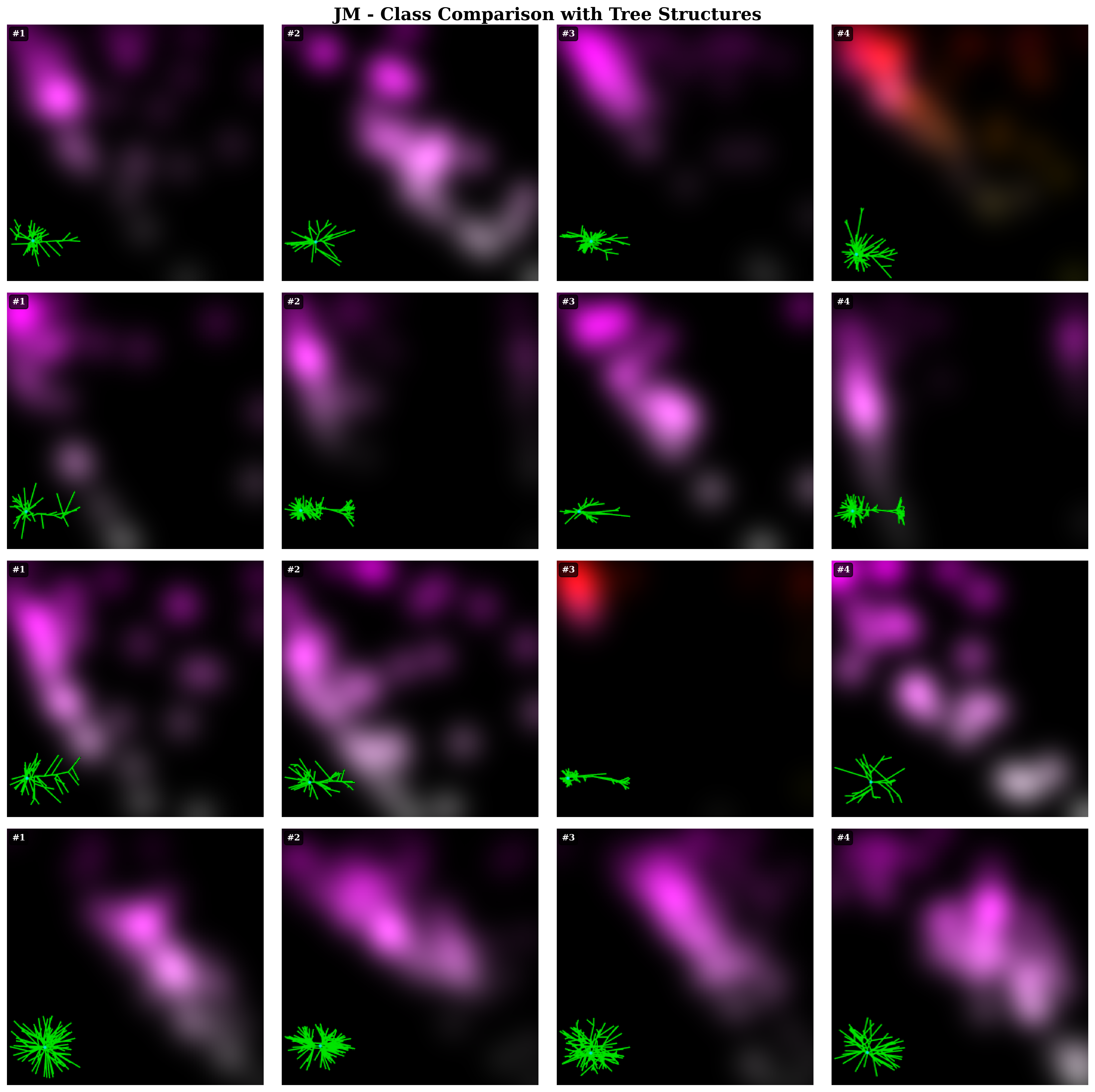}
    \includegraphics[width=0.32\linewidth]{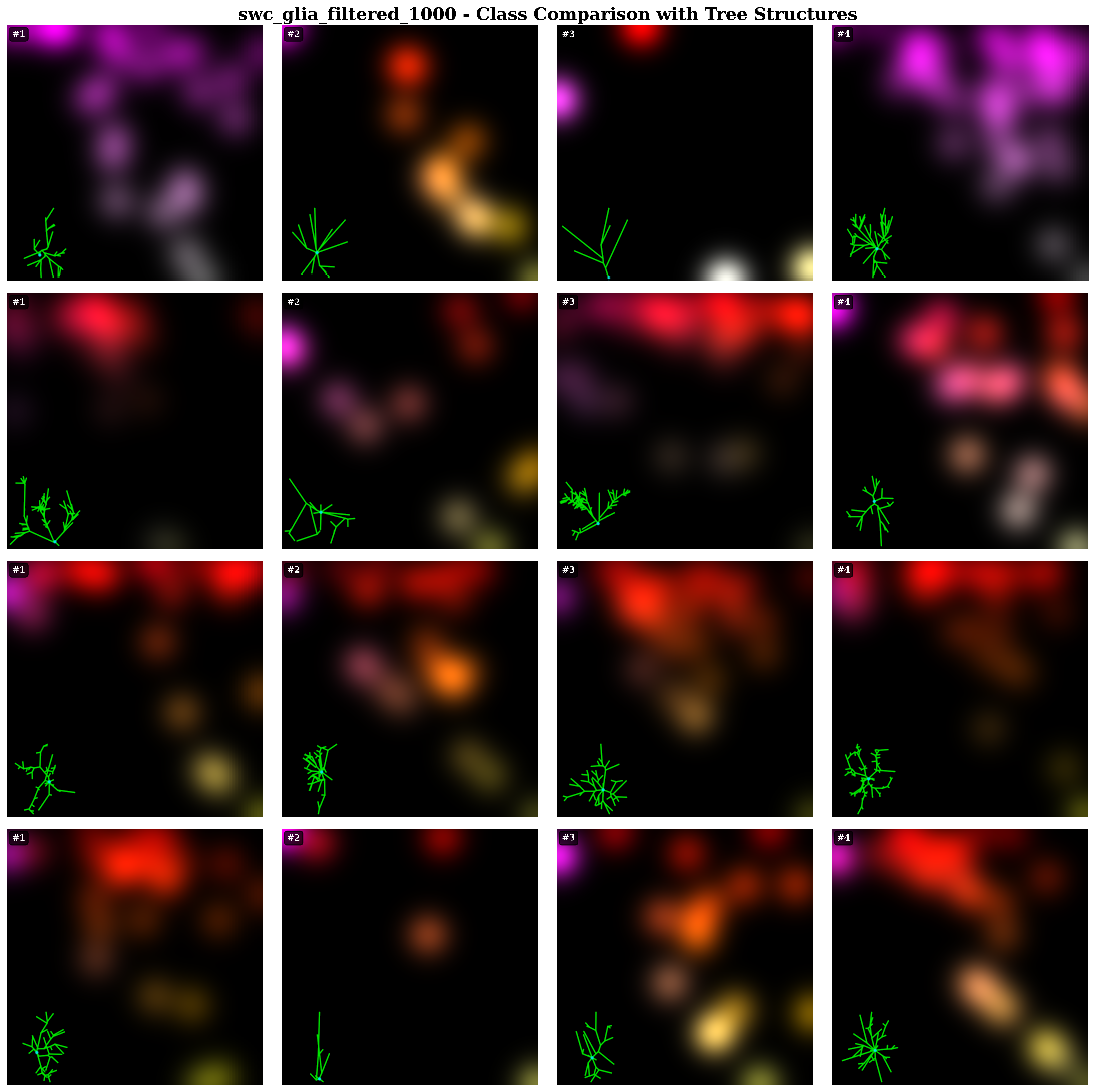}
    \caption{Representative examples showing tree graphs overlaid on their persistence images across different morphological classes. \textbf{Left}: ACT dataset (4 cortical layers). \textbf{Center}: JML-4 dataset (cortical layers and thalamic VPM neurons). \textbf{Right}: Glia dataset (4 species). Each row represents a distinct class, with 4 samples per class. The green tree structures illustrate how branching topology and spatial extent map onto the RGB persistence image encoding, where red captures unweighted density, green encodes persistence-weighted features, and blue represents radius-weighted information.}
    \label{fig:class_comparison}
\end{figure*}

To illustrate the relationship between tree graphs and their persistence image representations, Figure~\ref{fig:class_comparison} shows representative examples from ACT, JML, and Glia datasets with tree structures overlaid on their corresponding persistence images. Each row displays samples from a different morphological class, with the tree graph (green) superimposed on the multi-channel persistence image encoding.

The visualizations reveal how morphological features manifest in both representations. For ACT cortical neurons, layer-specific differences in dendritic complexity and branching patterns produce characteristic persistence signatures with varying radial extent (red channel intensity) and branch persistence (green channel). JML-4 samples demonstrate contrast between cortical projection neurons with complex dendritic arbors and thalamic VPM neurons with simpler, more compact morphologies. Glia morphologies demonstrate species-specific variations in process organization, reflected in the spatial distribution and intensity patterns across persistence channels.

The correspondence between tree complexity and persistence image structure confirms that our multi-channel encoding compresses topological information while preserving class-discriminative features. Within-class consistency (similar patterns across rows) and between-class variation (distinct signatures across rows) validate that both modalities capture biologically meaningful morphological differences used by GraPHFormer for classification.